\documentclass[runningheads]{llncs}

\usepackage{graphicx}

\usepackage{hyperref}
\usepackage{enumerate}   
\title{Linear time small coresets for $k$-mean clustering of segments with applications~\thanks{First published in WALCOM 2026 by Springer Nature}} 

\newcommand{\semipart}[1]{%
	\clearpage
	\thispagestyle{plain} % or empty if you don’t want headers/footers
	\begin{center}
		\vspace*{0.3\textheight} % vertical centering
		{\Huge\bfseries #1\par} % big bold title
		\vfill
	\end{center}
	\clearpage
}

\usepackage[table]{xcolor}
\usepackage[noend,ruled,vlined,linesnumbered]{algorithm2e}

\newcommand{\our}{\textsc{Our-Method}}

\newcommand{\LineClust}{\textsc{Line-Clustering}}

\newcommand{\OneSegmentCoreset}{\textsc{Seg-Coreset}}
\newcommand{\Coreset}{\textsc{Coreset}}
\newcommand{\ConvexCoreset}{\textsc{Convex-Coreset}}

\newcommand{\of}[1]{\left(#1\right)}
\newcommand{\abs}[1]{\left| #1\right|}
\newcommand{\loss}{\mathrm{loss}}

\newcommand{\ball}{\mathrm{ball}}

\newcommand{\lip}{\mathrm{lip}}
\newcommand{\eps}{\varepsilon}
\newcommand{\dist}{D}
\newcommand{\Q}{\mathcal{Q}}
\newcommand{\REAL}{\ensuremath{\mathbb{R}}}

\newcommand{\br}[1]{\left\{#1\right\}}

\usepackage{amsmath}
\usepackage{amsfonts}

\DeclareMathOperator*{\argmin}{arg\,min}

\usepackage{graphicx}
\usepackage{enumerate}

\author{David Denisov \inst{1} \and
	Shlomi Dolev\inst{2} \and
	Dan Feldman\inst{1} \and
	Michael Segal\inst{2}
}
\authorrunning{Denisov. et al.}
\titlerunning{Coreset for $k$-mean clustering of segments}
\institute{University of Haifa, Haifa, Israel.\\
	E-mail: David Denisov: \email{daviddenisovphd@gmail.com}. \and
	Ben-Gurion University of the Negev, Beer-Sheva, Israel}

\date{September 21, 2025}

\begin{document}
	
	\maketitle
	
	\begin{abstract}
	We study the $k$-means problem for a set $\mathcal{S} \subseteq \mathbb{R}^d$ of $n$ segments, aiming to find $k$ centers $X \subseteq \mathbb{R}^d$ that minimize 
	$D(\mathcal{S},X) := \sum_{S \in \mathcal{S}} \min_{x \in X} D(S,x)$, where $D(S,x) := \int_{p \in S} |p - x| dp$
	measures the total distance from each point along a segment to a center. Variants of this problem include handling outliers, employing alternative distance functions such as M-estimators, weighting distances to achieve balanced clustering, or enforcing unique cluster assignments. For any $\varepsilon > 0$, an $\varepsilon$-coreset is a weighted subset $C \subseteq \mathbb{R}^d$ that approximates $D(\mathcal{S},X)$ within a factor of $1 \pm \varepsilon$ for any set of $k$ centers, enabling efficient streaming, distributed, or parallel computation. We propose the first coreset construction that provably handles arbitrary input segments. For constant $k$ and $\varepsilon$, it produces a coreset of size $O(\log^2 n)$ computable in $O(nd)$ time. Experiments, including a real-time video tracking application, demonstrate substantial speedups with minimal loss in clustering accuracy, confirming both the practical efficiency and theoretical guarantees of our method.
		
	\end{abstract}
	
	\section{Introduction} \label{sec: Introduction}
	Segment clustering generalizes point and line clustering and arises in applications such as video tracking, facility location, and spatial data analysis. While point clustering is well-studied~\cite{kmeans++} and line clustering has a few provable approximations~\cite{YairKClustering}, no provable algorithms exist for segments. Segments are more challenging because they represent continuous sets of points, making the loss function an integral rather than a sum.
	Nevertheless, segments naturally model real-world structures such as maps, road networks, and image features.
	Efficient segment clustering enables the scalable analysis of such structured data, supporting tasks such as object tracking, pattern detection, and optimization over vectorized inputs. In this work, we introduce novel compression and approximation algorithms for segment clustering that provide both theoretical guarantees and practical performance, extending classical point-based methods to structured, high-dimensional inputs.
	Applications of our methods are demonstrated in Section~\ref{sec: video application} through a real-time video tracking system leveraging segment coresets for fast and accurate processing.

	\subsection{Segment clustering} \label{sec: Segment clustering}
	To formalize the problem, we introduce the following definitions and notations.
	\begin{definition} \label{def: segment}
		For two given vectors $u,v$ in $ \REAL^d$, a \emph{segment} is a function $\ell:[0,1]\to \REAL^d$ that is defined by $\ell(x) = u + v x$, for every $x\in [0,1]$.
		For every pair $(p,p') \in \REAL^d \times \REAL^d$ of points, let $\dist(p,p'):=\|p-p'\|_2$ denote the Euclidean distance between $p$ and $p'$.
	\end{definition}
	
	Formally, we generalize the classical $k$-means problem to segments, defining the segment clustering problem as follows.
	
\begin{figure}[h]
		\begin{center}
			\includegraphics[width=0.495\textwidth]{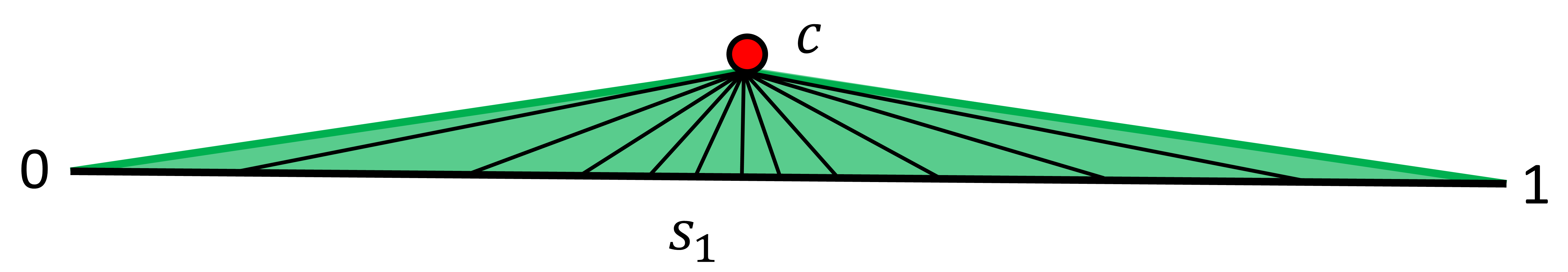}
			\includegraphics[width=0.495\textwidth]{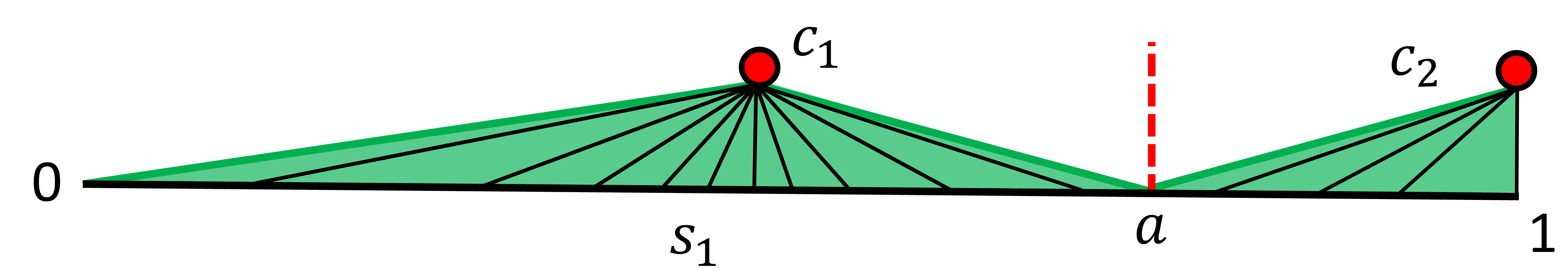}
			\caption
			{Visual illustration of Problem~\ref{problem 1}. 
				Left: the loss of a single center $c\in\mathbb{R}^2$ to a segment $s_1$, with the green area representing the integrated distance.
				Right: the loss for two centers $C=\{c_1,c_2\}$, where the green area is split at $a\in(0,1)$, assigning points closer to $c_1$ or $c_2$ accordingly.
			}
			\label{fig: loss ill}
		\end{center}
	\end{figure}

	\begin{problem}[Segment clustering]\label{problem 1}
		Let $S$ be a set of $n$ segments in $\REAL^d$, and let $k\geq1$ be an integer.
		Let $\Q$ be the set of all pairs $(C,w)$, where $C$ is a set of $|C|=k$ points in $\REAL^d$ called \emph{centers}, and $w: C \to [0,\infty)$ is a weight function representing the influence of each center.
		The \emph{$k$-segment mean} of $S$, which we aim to find, is a set $(C,w)\in \Q$ that minimizes
		\begin{equation}\label{eq: problem main}
			\loss(C,w) := \sum_{s\in S} \int_0^1 \min_{c \in C} w(c) \dist\big(c,s(x)\big) dx.
		\end{equation}
	\end{problem}
	The integral exists due to the continuity of the Euclidean distance.
	The corresponding loss $\loss$ is illustrated in Fig.~\ref{fig: loss ill}.
	
	In this paper, we aim to derive a provable approximation to Problem~\ref{problem 1}, i.e., the $k$-segment mean, which has broad applicability as demonstrated in our experiments. Our results naturally extend to the setting where each segment is assigned to a single cluster. The compression remains the same, but solving the problem over the compressed set corresponds to set clustering~\cite{Sets-Clustering}, which is less straightforward than $k$-means; in contrast, $k$-means (e.g., via k-means++~\cite{kmeans++}) is efficiently implemented in many standard libraries.
	
	\subsection{Coresets} \label{sec:coresets}
	Informally, for an input set $L$ of segments, a coreset $C$ is a (weighted) set of points that approximates the loss $\loss(L,(Q,w))$ for any weighted set $(Q,w)$ of size $k$ up to a factor of $1\pm\varepsilon$, with computation time depending only on $|C|$. Ideally, $C$ is much smaller than $L$, making it efficient. For more on the motivation behind coresets, see “Why coresets?” in~\cite{feldman2020core,dan_kdd}.
	
	\textbf{Coreset for segments.}
	The segment clustering problems from Sections~\ref{sec: Segment clustering} and~\ref{Problem definition} admit a small coreset, a weighted subset of points sampled from the segments. While many coresets are subsets of the input, there are exceptions~\cite{jubran2021coresets,k-Segmentation}, including ours, which is a weighted set of points rather than the original segments. There is extensive work on fitting weighted centers to weighted points~\cite{outliers-resistance,kmeans++}, but little on fitting points to segments or handling non-discrete integrals. Thus, a coreset represented as weighted points provides a practical and usable solution for segment clustering.
	
	\textbf{Coreset for convex sets and hyperplane fitting.}
	Convex sets can be seen as unions of infinitely many segments, which allows us to construct small coresets for $n\geq 1$ convex sets (Algorithm~\ref{algorithm - convex coreset}). Moreover, our focus on fitting weighted centers naturally extends the approach to hyperplanes~\cite{outliers-resistance}.
	\textbf{Relation to our goal.}
	We aim to derive a provable approximation to the $k$-segment mean. Given the extensive work on fitting weighted centers to weighted points, we compute a coreset as a weighted set of points. Any provable or optimal solution over this coreset directly yields an approximation to the original problem. A formal statement appears in Corollary~\ref{ptas}.
	\subsection{Novelty}
	Our main contribution is a provable reduction of the segment clustering problem to point clustering, yielding a coreset construction via prior work~\cite{k_means_code,outliers-resistance}.
	For a function $f:\mathbb{R}\to\mathbb{R}$ and integer $n\ge 1$, we call $\sum_{i=1}^n f(i/n)$ an \emph{$n$-discrete integral} of $f$ over $[0,1]$.
	
	\textbf{Discrete integrals.}
	Previous work~\cite{integration_coreset} shows that a weighted subset of size $O(\log n/\varepsilon^2)$ approximates the $n$-discrete integral for certain functions. We generalize this to the continuous integral $\int_0^1 f(x) dx$, yielding the first provably small coreset for this setting. Applying~\cite{integration_coreset} directly to a segment would yield an infinite coreset, whereas Lemma~\ref{lemma: deterministic construction} provides a finite one.
	
	\textbf{Riemann sums.}
	Our approach is related to Riemann sums~\cite{Riemann_orig}, but unlike prior work, we provide a hard bound for finite samples (Theorem~\ref{coreset theorem}).
	
	\textbf{Relation to prior coreset results.}
	Theorem~\ref{coreset theorem} generalizes previous coreset constructions for points~\cite{outliers-resistance} to handle infinitely many points along segments. 
	Similarly, while~\cite{YairKClustering} constructs coresets for lines. However, their approach does not trivially extend to segments.
	\subsection*{Paper structure}
	Due to space limitations, most proofs and additional experiments are in the appendix. Section~\ref{Theoretical results} presents our main theoretical results and algorithms. Section~\ref{sec:ER} compares our approach to previous line-clustering work~\cite{YairKClustering}, and Section~\ref{sec: video application} demonstrates a video tracking and detection method based on our results. In the appendix, Section~\ref{sec: sen bound} bounds the sensitivity of a single $r$-Lipschitz function (Lemma~\ref{mini sen bound}), Section~\ref{sec: sen minimum over} bounds the sensitivity of a function defined as the minimum over such functions (Lemma~\ref{general sen bound}), Section~\ref{sec: det core} proves a deterministic coreset construction given a sensitivity bound, Section~\ref{sec: alg 1 proof} proves the properties of Algorithm~\ref{algorithm - deterministic sample}, and Section~\ref{sec: Algorithm 3} generalizes the results to multi-dimensional shapes.
	
	\section{Theoretical Results}\label{Theoretical results}
	In the following section, we present our main theoretical results, which essentially outline an efficient data reduction scheme for Problem~\ref{problem 1} defined in the previous section.
 	Note that the loss function we define supports cluster weighting and a distance function as in~\cite{outliers-resistance}, which allows immediate generalization of the support for M-estimators, outlier support, and more from~\cite{outliers-resistance}.
	\subsection{Notation and Definitions}  \label{Problem definition}
	\textbf{Notations. }
	Throughout this paper, we assume that $k,d\geq 1$ are integers and denote by $\REAL^d$ the set of all $d$-dimensional real vectors.
	A \emph{weighted set} is a pair $(P,w)$ where $P$ is a finite set of points in $\REAL^d$ and $w: P \to [0,\infty)$ is a \emph{weight function}.
	For simplicity, we denote $\log(x):=\log_2(x)$.
	\begin{definition}  \label{def: log-Lipschitz}
		Let $r\geq 0$, and let $h: \REAL \to [0,\infty)$ be a non-decreasing function.
		We say that $h$ is \emph{$r$-log-Lipschitz} iff, for every $c \ge 1$ and $x \ge 0$, we have $h(cx) \leq c^r h(x)$.
	\end{definition}
	\begin{definition} \label{def: symmetric-$r$ function}
		A \emph{symmetric-$r$} function is a function $f: \REAL\to \REAL$, such that there is $a \in \REAL$ and an $r$-log-Lipschitz function $\tilde{f}:[0,\infty) \to [0,\infty)$, that for every $x\in \REAL$ satisfies $f(x) = \tilde{f}(|x-a|)$.
	\end{definition}
	
	In the following definitions we define $\lip$, a global $r$-log-Lipschitz function, with $r,t,d^*\in [0,\infty)$, and $\dist$, a distance function; those definitions are inspired by~\cite{outliers-resistance}.
	\begin{definition}[\textbf{global parameters}]\label{def: global-lip}
		Let $\lip:\REAL \to \REAL$ be a symmetric-$r$ function for some $r\geq 0$, and suppose that for every $x\in [0,\infty)$ we can compute $\lip(x)$ in $t$ time, for some integer $t\geq 1$; see Definition~\ref{def: log-Lipschitz}.
		For every weighted set $(C,w)$ of size $|C|=k$ and $p\in \REAL^d$ let 
		$\displaystyle \dist((C,w),p):=\min_{c\in C} \lip \big(w(c)\dist(c,p)\big).$
	\end{definition}
	\begin{definition}[\textbf{VC-dimension}]\label{def: vc-dim}
		For every $P\subset \REAL^d,r\geq 0$ and any weighted set $(C,w)$ of size $k$ let 
		\[
		\ball(P,(C,w),r):= \br{p\in P \mid \dist\big((C,w),p\big) \leq r}.
		\]
		Let $B:=\br{\ball(P,(C,w),r) \mid \big(P,r,(C,w)\big)\in \tilde{\Q}}$, where $\tilde{\Q}$ is the union over every $P\subset \REAL^d,r\geq 0$ and any weighted set $(C,w)$ of size $k$.
		Let $d^*$ denote the VC-dimension associated with $\lip$, which is the smallest positive integer such that for every finite $S\subset \REAL^d$ we have
		\[
		\Big| \big\{(S\cap \beta) \mid \beta\in B \big\} \Big| \leq |S|^{d^*}.
		\]
	\end{definition}
	We emphasize that the choice of the distance function $D$ dictates the appropriate values of $r,t,d^*$.
	
	Utilizing this definition, we define an algorithm, as stated in the following theorem, which follows from Theorem 5.1 of~\cite{outliers-resistance}.
	\begin{theorem}\label{th: outliers-resistance coreset}
		Let $k'\in (k+1)^{O\of{k}}$.
		There is an algorithm $\textsc{Core-set}(P,k,\eps,\delta)$ that gets:
		\textbf{(i)}  A set $P$ of points in $\REAL^d$.
		\textbf{(ii)} An integer $k\geq 1$.
		\textbf{(iii)} Input parameters $\eps,\delta\in (0,1/10)$,
		such that its output $(S,w):=\textsc{Core-set}(P,k,\eps,\delta)$ is a weighted set satisfying 
		Claims~(i)--(iii) as follows:
		\begin{enumerate}[(i)]
			\item  With probability at least $1-\delta$, for every weighted set $(C,w)$ of size $|C|=k$, we have
			\[\left| \sum_{p\in P} \dist\big((C,w'),p\big) - \sum_{p\in S} \Big(w(p) \dist\big((C,w'),p\big) \Big) \right|
			\leq \eps \cdot \sum_{p\in P} \dist\big((C,w'),p\big).
			\]
			\item $\displaystyle |S| \in \frac{k' \cdot \log^2 n }{\eps^2} \cdot O\of{d^* +\log\of{\frac{1}{\delta}}}$.
			\item The computation time of a call to $\textsc{Core-set}(P,k,\eps,\delta)$ is in 
			\[
			n t k' + t k' \log(n) \cdot \log\big(\log(n)/\delta\big)^2 
			+ \frac{k' \log^3 n}{\eps^2} \cdot \left(d^* + \log \of{\frac{1}{\delta}} \right).
			\]
		\end{enumerate}
	\end{theorem}

	\begin{definition}[Loss function]\label{def: loss function}
		Let $\ell:[0,1]\to \REAL^d$ be a segment; see Definition~\ref{def: segment}.
		We define the \emph{fitting loss} of a weighted set $(C,w)$ of size $k$, as
		\begin{equation*}
			\loss\big((C,w),\ell\big) = 
			\int_0^1 \dist\big((C,w),\ell(x)\big) dx.
		\end{equation*}
		Given a finite set $L$ of segments and  a weighted set $(C,w)$ of size $|C|=k$, we define \emph{the loss of fitting} $(C,w)$ to $L$ as
		\begin{equation} \label{eq: loss l2s}
			\loss\big((C,w),L\big) = \sum_{\ell \in L} \loss\big((C,w),\ell\big).
		\end{equation}
		Given such a set $L$,  the goal is to recover a weighted set $(C,w)$ of size $k$ that minimizes $\loss\big((C,w),\ell\big)$.
	\end{definition}

	\begin{definition} [$(\eps,k)$-coreset] \label{def: eps coreset}
		Let $\ell$ be a segment, and let $\eps>0$ be an error parameter; see Definition~\ref{def: segment}.
		A weighted set $(S,w)$ is an \emph{$(\eps,k)$-coreset} for $\ell$ if for every weighted set $Q:=(C,w)$ of size $|C|=k$ we have 
		$ \displaystyle 
		\abs{\loss(Q,\ell) - \sum_{p\in S} w(p) \dist(Q,p)} \leq \eps \cdot \loss(Q,\ell).
		$
		More generally a weighted set $(S,w)$ is an \emph{$(\eps,k)$-coreset} for a set $L$ of segments if for every weighted set $Q:=(C,w)$ of size $|C|=k$ we have 
		\[
		\abs{\loss(Q,L) - \sum_{p\in S} w(p)\cdot \dist(Q,p)} \leq \eps \cdot \loss(Q,L).
		\]
	\end{definition}

	\subsection{Algorithms}\label{sec: Algorithms}
	\subsubsection{Algorithm~\ref{algorithm - deterministic sample}.}\label{sec: Algorithm 1}
	Algorithm~\ref{algorithm - deterministic sample} and its corresponding theorem (Theorem~\ref{deterministic sample theorem}) are our main theoretical results. The algorithm provides a simple deterministic coreset for the single-segment case ($n=1$) in Definition~\ref{def: loss function} (Eq.~\eqref{eq: loss l2s}). The parameter $r$ depends on the distance function $D$ (e.g., $r=1$ for absolute error, $r=2$ for MSE). Its novelty lies in bounding the contribution of each point on the segment to the loss, which is constant. While a uniform sample could approximate this at the cost of failure probability and larger size, we use a deterministic construction generalizing~\cite{k-Segmentation}. 
	We emphasize that the algorithm is not a heuristic but comes with provable guarantees. Although simple in form, it relies on non-trivial analysis, which enables this very simplicity.
	
	\begin{algorithm}[h]
		\caption{$\OneSegmentCoreset(\ell,k,\eps)$; see Theorem~\ref{deterministic sample theorem}.\label{algorithm - deterministic sample}}
		\SetKwInOut{Input}{Input}
		\SetKwInOut{Output}{Output}
		\Input{A segment $\ell: [0,1] \to \REAL^d$, an integer $k\geq 1$, and error $\eps\in (0,1/10]$; see Definition~\ref{def: segment}.}
		\Output{A weighted set $(S,w)$, which is an $(\eps,k)$-coreset of $\ell$; see Definition~\ref{def: eps coreset}.}
		
		$\displaystyle \eps':= \left\lceil \frac{4k\cdot (20k)^{r+1}}{\eps}\right\rceil$ 
		\tcp{$r$ is as defined in Definition~\ref{def: global-lip}.}
		
		$S:= \br{ \ell(i/\eps') \mid i\in \br{0,\cdots,\eps'}}$ \tcp{note that $\displaystyle \eps' \in \frac{O(k)^{r+2}}{\eps}$.}
		
		Let $w(p) := 1/\eps'$ for all $p \in S$.
		
		\Return $(S,w)$.
	\end{algorithm}
	
	The following theorem states the desired properties of Algorithm~\ref{algorithm - deterministic sample}; see Theorem~\ref{deterministic sample theorem proof} for its proof.
	\begin{theorem}\label{deterministic sample theorem}
		Let $\ell: [0,1] \to \REAL^d$ be a segment, and let $\eps \in (0,1/10]$; see Definition~\ref{def: segment}.
		Let $(S,w)$ be the output of a call to $\OneSegmentCoreset(\ell,k,\eps)$; see Algorithm~\ref{algorithm - deterministic sample}. 
		Then $(S,w)$ is an $(\eps,k)$-coreset for $\ell$; see Definition~\ref{def: eps coreset}.
	\end{theorem}
	Setting $\tilde{f}(x) = x^2$ and using previous results, we obtain a provable $k$-means approximation for segments by constructing coresets per segment (Theorem~\ref{deterministic sample theorem}) and applying $O(\log n)$ repetitions of~\cite{kmeans++} on their union.
	\begin{corollary}\label{ptas}
		Let $L$ be a set of $n$ segments, and let $\eps,\delta \in (0,1/10]$.
		We can compute in $O(n k^3\eps \log(1/\delta))$ time, a set $P\subset \REAL^d,|P|=k$ such that with probability at least $1-\delta$ we have 
		\[
		\loss(P,L) \in O(\log k) \min_{P'\in \REAL^d,|P'|=k} \loss(P',L).
		\]
	\end{corollary}
	
	\subsubsection{Algorithm~\ref{algorithm - coreset}.}\label{sec: Algorithm 2}
	We use Algorithm~\ref{algorithm - deterministic sample} to convert each segment into points, then apply the compression scheme from~\cite{outliers-resistance} on their union.
	
	\textbf{Overview of Algorithm~\ref{algorithm - coreset}.}
	Given a set $L$ of $n$ segments, an integer $k\ge 1$, and parameters $(\varepsilon,\delta)\in(0,1/10]$, the algorithm outputs a weighted set $(P',w')$ that, with probability at least $1-\delta$, is an $(\varepsilon,k)$-coreset for $L$.
	Each segment is first reduced to points using Algorithm~\ref{algorithm - deterministic sample}, producing a weighted set. The union of these sets is then processed via $\textsc{Core-set}$ (Theorem~\ref{th: outliers-resistance coreset}) to further reduce the coreset size.

	\begin{algorithm}[h]
		\caption{$\Coreset(L,k,\eps,\delta)$; see Theorem~\ref{coreset theorem}.\label{algorithm - coreset}}
		\SetKwInOut{Input}{Input}
		\SetKwInOut{Output}{Output}
		
		\Input{A finite set $L$ of segments, an integer $k\geq1$, and input parameters $\eps,\delta\in (0,1/10]$.}
		\Output{A weighted set $(P,w)$, which, with probability at least $1-\delta$, is an $(\eps,k)$-coreset of $L$; see Definition~\ref{def: eps coreset}.}
		
		For every $\ell \in L$ let $(P'_{\ell},w_\ell):=\OneSegmentCoreset(\ell,k,\eps/2)$ \tcp{see Algorithm~\ref{algorithm - deterministic sample}, the division by $2$ accounts for the output being further reduced.}
		
		$\displaystyle P':= \bigcup_{\ell\in L} P'_{\ell}$ 
		
		$(P,w'):=\textsc{Core-set}(P',k,\eps/4,\delta)$\tcp{see Theorem~\ref{th: outliers-resistance coreset}.}
		
		$\displaystyle \eps':= \left\lceil  \frac{8k\cdot (20k)^{r+1}}{\eps} \right\rceil$ \tcp{$r$ is as defined in Definition~\ref{def: global-lip}.}
		
		Set $w(p) := w'(p)/\eps'$ for every $p\in P'$.
		
		\Return $(P,w)$.
	\end{algorithm}

\begin{figure}[h]
		\centering
		\includegraphics[width=0.328\textwidth]{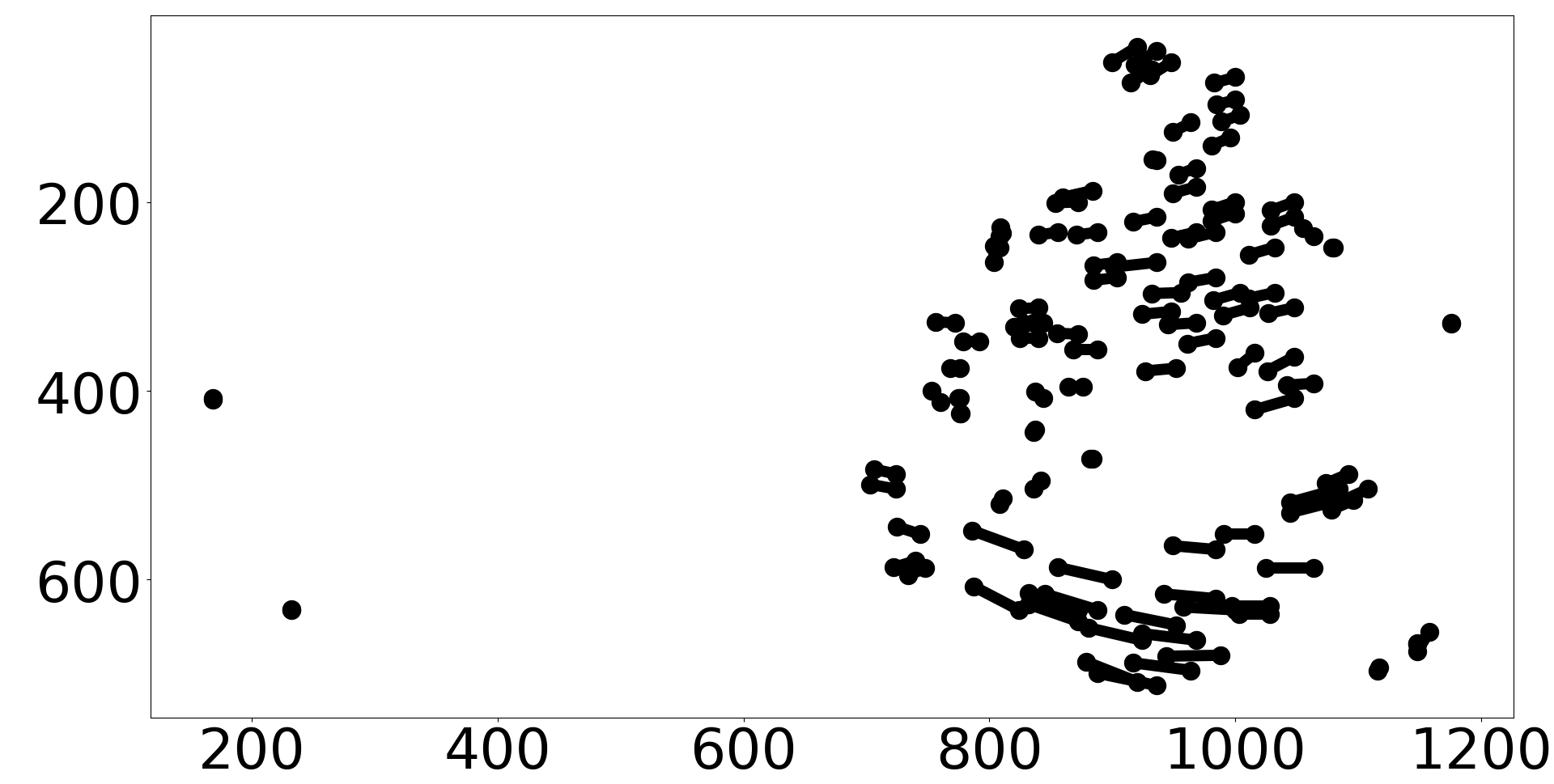}
		\includegraphics[width=0.328\textwidth]{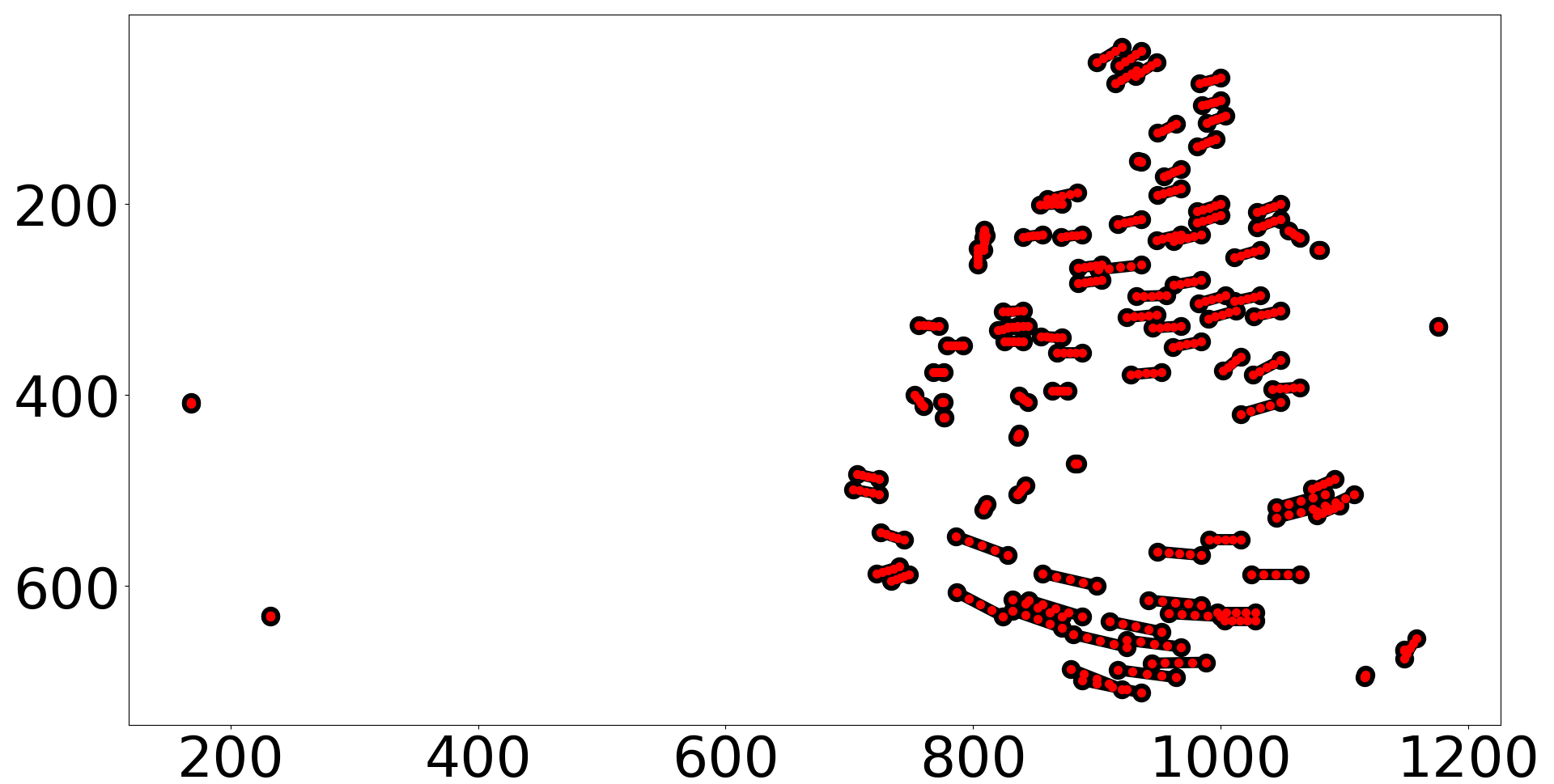}
		\includegraphics[width=0.328\textwidth]{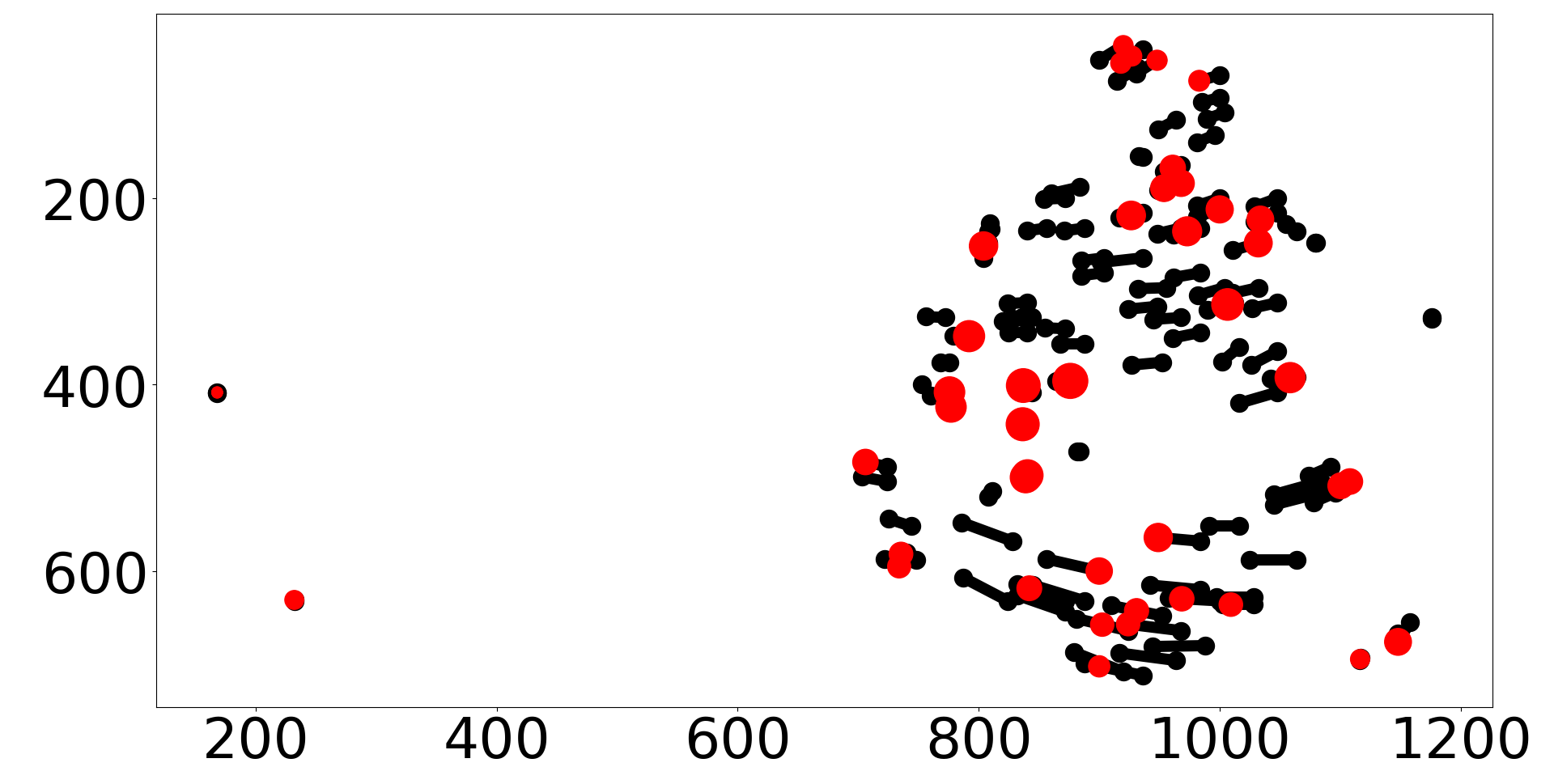}
		\caption{Illustration of Algorithm~\ref{algorithm - coreset}. 
			\textit{Left:} The input $n=100$ segments, obtained from the motion vectors of the video~\cite{Bunny_video}.
			\textit{Middle:} The union of the outputs of the calls to Algorithm~\ref{algorithm - deterministic sample} computed at Line 1 in the call to Algorithm~\ref{algorithm - coreset} is the red points (all with equal weight) and black lines representing the input segments.
			\textit{Right:}  The final coreset returned by Algorithm~\ref{algorithm - coreset}, which is obtained by further sampling from the outputs of Algorithm~\ref{algorithm - deterministic sample} via \cite{k_means_code}.
			Here, the size of each point is proportional to its weight.
		}
		\label{Fig: - core_example}
	\end{figure}

	The structure of Algorithm~\ref{algorithm - coreset} is illustrated in Figure~\ref{Fig: - core_example}; note that we have substituted $\textsc{Core-set}$ in Definition~\ref{def: global-lip} by~\cite{k_means_code}.
	For an explanation on motion vectors, see Section~\ref{sec: motion vectors explenation}.

	The following theorem states the desired properties of Algorithm~\ref{algorithm - coreset}, for its proof see Theorem~\ref{coreset theorem proof}.
	\begin{theorem} \label{coreset theorem}
		Let $L$ be a set of $n$ segments, and let $\eps,\delta \in (0,1/10]$.
		Let $\displaystyle m:=\frac{8k n \cdot (20k)^{r+1}}{\eps}$ and $k'\in (k+1)^{O(k)}$.
		Let $(P,w)$ be the output of a call to $\Coreset(L,k,\eps,\delta)$; see Algorithm~\ref{algorithm - coreset}.
		Then Claims~(i)--(iii) hold as follows:
		\begin{enumerate}[(i)]
			\item With probability at least $1-\delta$, we have that $(P,w)$ is an $(\eps,k)$-coreset of $L$; see Definition~\ref{def: eps coreset}.
			\item $\displaystyle |P| \in \frac{k' \cdot \log^2 m }{\eps^2} \cdot O\of{d^* +\log\of{\frac{1}{\delta}}}$.
			\item The $(\eps,k)$-coreset $(P,w)$ can be computed in time
			\[
			m t k' + t k' \log(m) \cdot \log^2\big(\log(m)/\delta\big) 
			+ \frac{k' \log^3 m}{\eps^2} \cdot \left(d^* + \log \of{\frac{1}{\delta}} \right).
			\]
		\end{enumerate}
	\end{theorem}

	\section{Empirical Evaluation} \label{sec:ER}
	We evaluate the quality of our approximation against the line-clustering method from~\cite{YairKClustering}.
	For consistency, we measure error using the sum of squared distances (corresponding to the Gaussian noise model~\cite{Gaussian_noise}), which allows reduction of the task to the standard $k$-means, solved with the $k$-means++ approximation~\cite{kmeans++}.
	Accordingly, we apply only Algorithm~\ref{algorithm - deterministic sample} to reduce each segment to a small set of representative points (coreset).
	Our implementation (Python 3.8, NumPy) is tested on both synthetic and real datasets, with open-source code available in~\cite{opencode}.
	
	\textbf{Methods.} 
	We compare:\\
	\textbf{$\our$.} For each segment, compute a coreset of 10 points using \\$\OneSegmentCoreset$, then apply $k$-means++~\cite{kmeans++} to cluster the resulting points.\\
	\textbf{$\LineClust$.}  Line-clustering from~\cite{YairKClustering}, applied to segments extended to lines.
	
	\textbf{Note that $\LineClust$ is designed for line-clustering rather than segment-clustering.}
	However, since~\cite{YairKClustering} evaluated their approach on road segments extended to infinite lines in the OpenStreetMap dataset~\cite{OpenStreetMap}, we include it here as the closest available baseline.
	To our knowledge, no provable or commonly used methods exist for segment-clustering, making this the only relevant comparison.
	
	\textbf{Loss.}
	Since the optimization problem is not elementary, we approximate it using the same coreset construction as our method, but with a larger size.
	For each segment $\ell \in L$, we compute $(P_\ell,w_\ell):=\OneSegmentCoreset(\ell,k,0.1)$ with $|P_\ell|=10{,}000$, a larger coreset used to obtain a more robust approximation, and define $P=\bigcup_{\ell \in L} P_\ell$.
	The loss is then $\sum_{p\in P}\min_{c\in C}\dist(p,c)^2/|P|$, i.e., the mean squared error (MSE).
	
	\textbf{Data.}
	We use three datasets for comparison (see Section~\ref{sec: demo} for details).
	\textbf{(i).} Synthetic segments in $\REAL^{10}$, where each endpoint is sampled uniformly from $[-1,1]^{10}$; the sample size varies across experiments.
	\textbf{(ii).} Motion vectors (see~\cite{h264}) from a short clip of~\cite{Bunny_video}, sampled independently, with probability proportional to segment length. This setting demonstrates applications in tracking and object detection (see Section~\ref{sec: bunny test}).
	\textbf{(iii).} Road segments from OpenStreetMap~\cite{OpenStreetMap} (via Geofabrik), using longitude/latitude coordinates and sampling proportional to segment length. We focus on the one million longest segments in Malaysia, Singapore, and Brunei, where fitted centers highlight key points in the road network relevant to facility location and infrastructure planning.
	
	\textbf{Hardware.} 
	We used a PC with an Intel Core i5-12400F, NVIDIA GTX 1660 SUPER (GPU), and 32GB of RAM.
	
	\textbf{Segment count.} $200$ to $1000$ in steps of $100$.
	
	\textbf{Repetitions.} Each experiment was repeated $40$ times.

\begin{figure}[h!]
		\centering
		\includegraphics[width=0.6\textwidth]{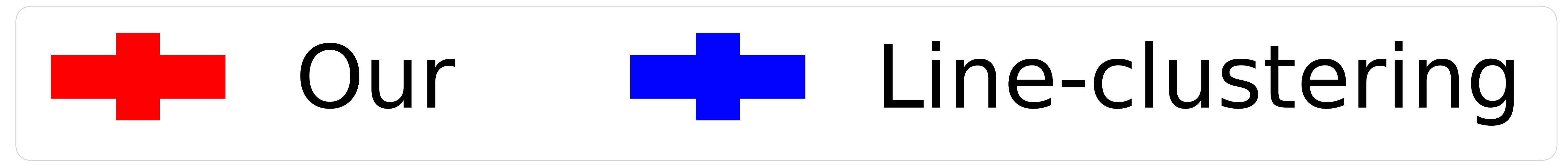}
		
		\includegraphics[width=0.325\textwidth]{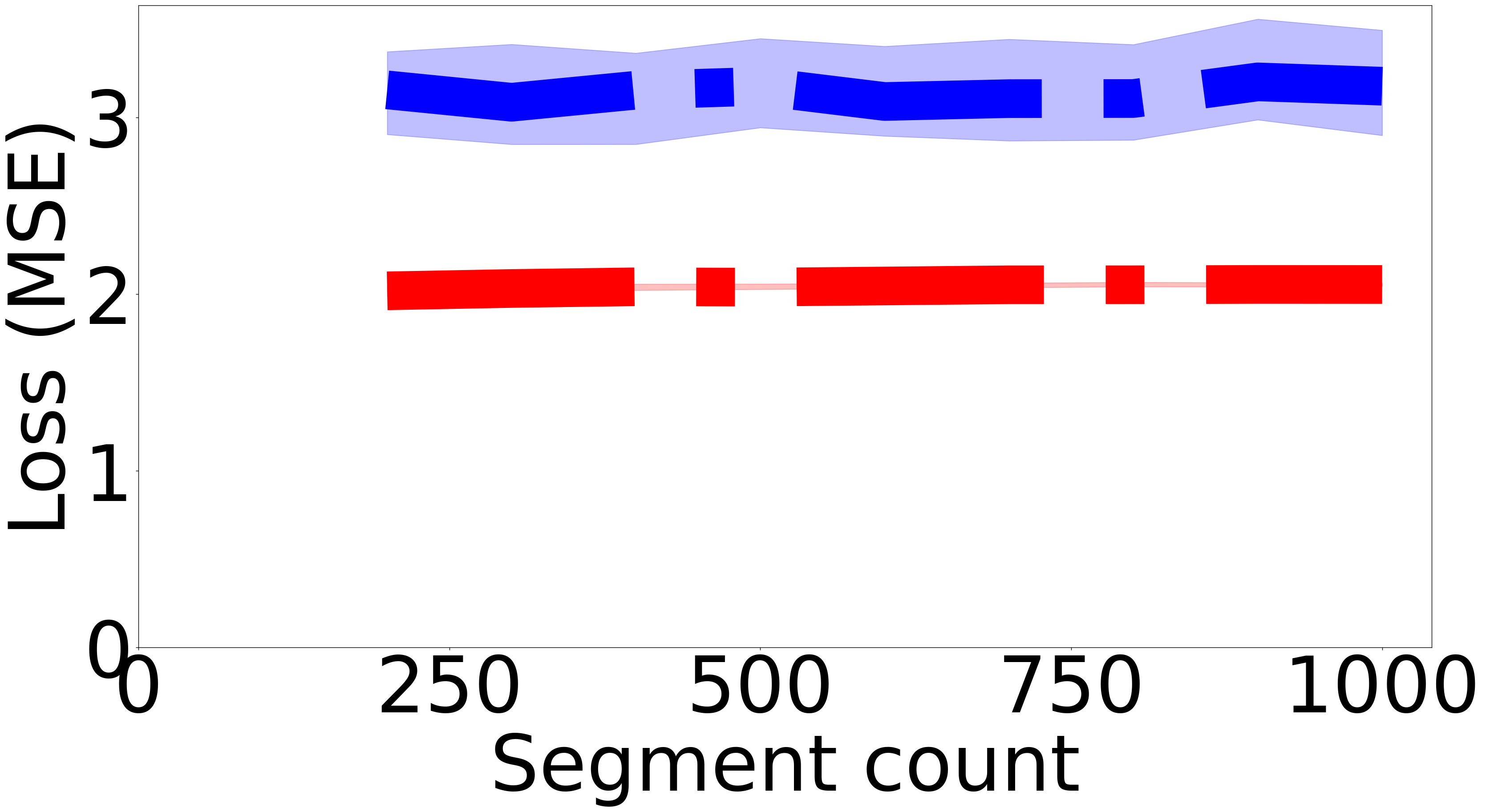}
		\includegraphics[width=0.325\textwidth]{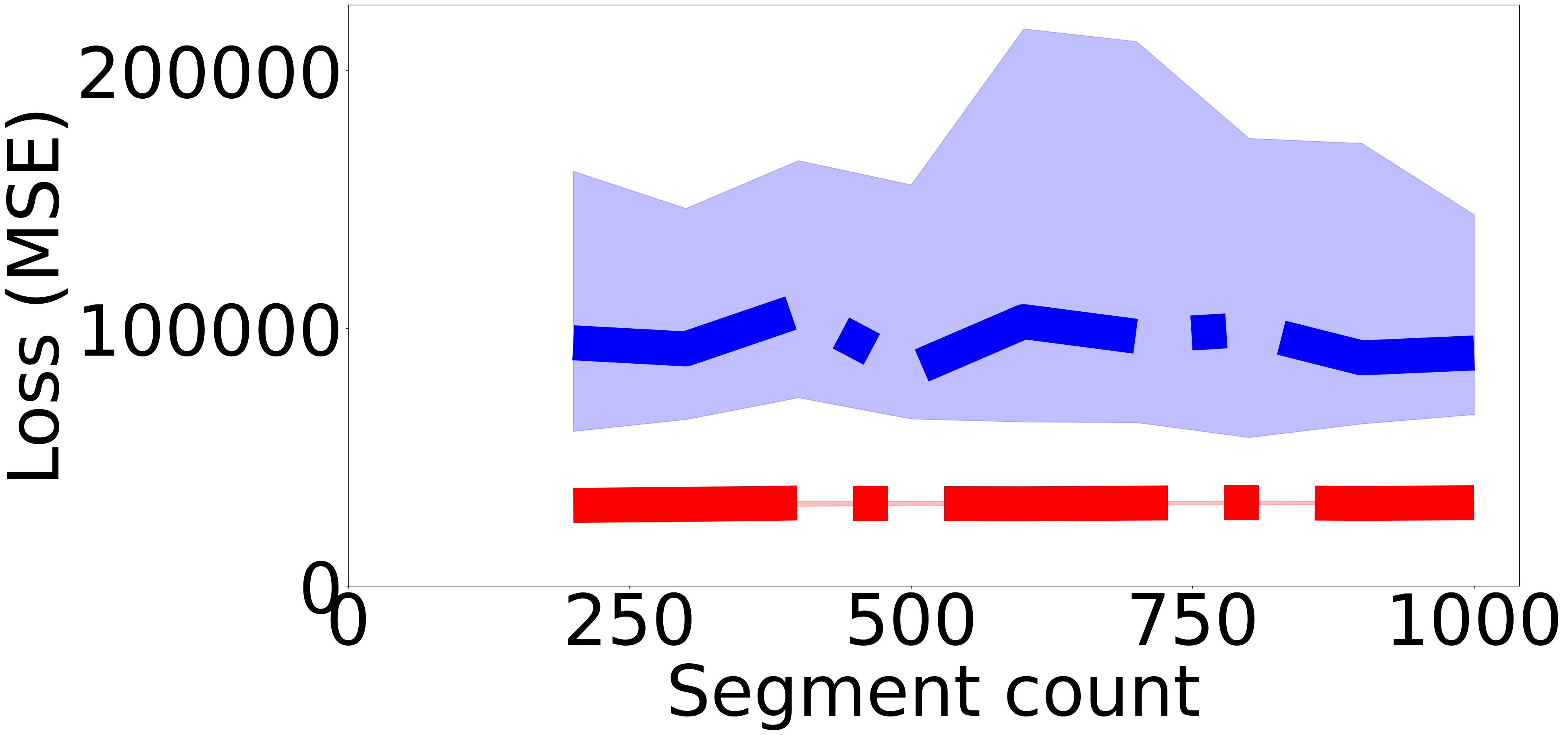}
		\includegraphics[width=0.325\textwidth]{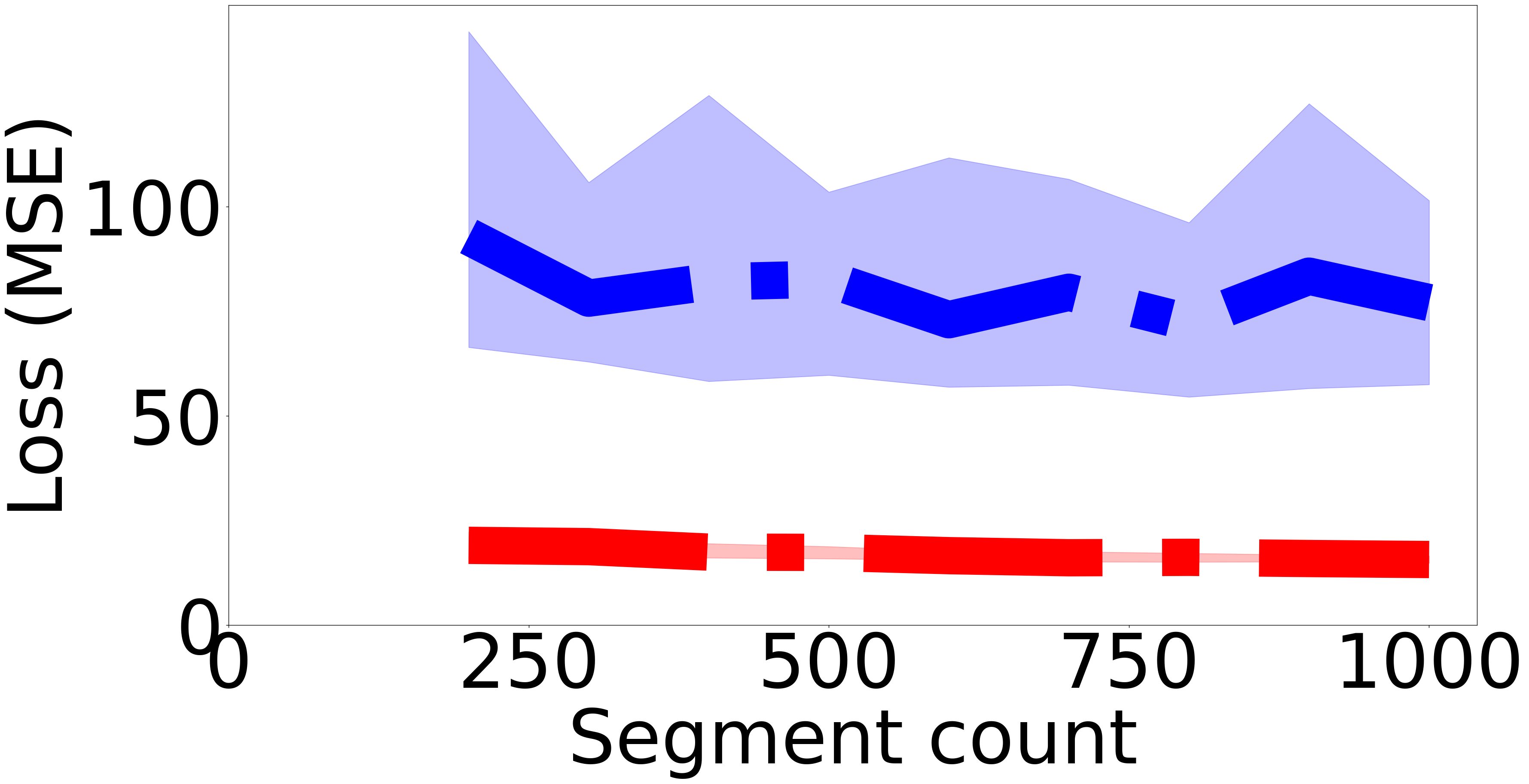}
		
		\includegraphics[width=0.325\textwidth]{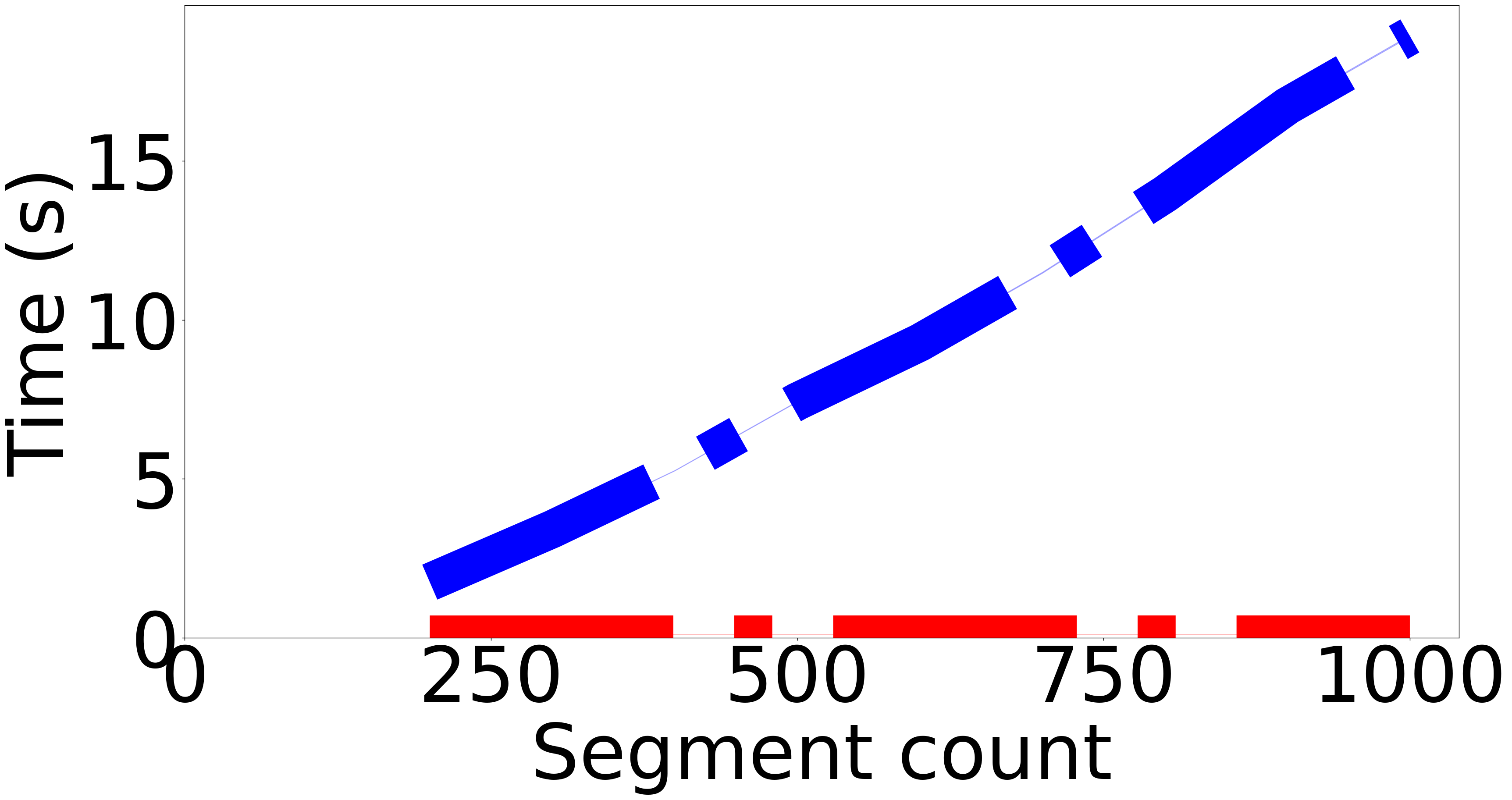}
		\includegraphics[width=0.325\textwidth]{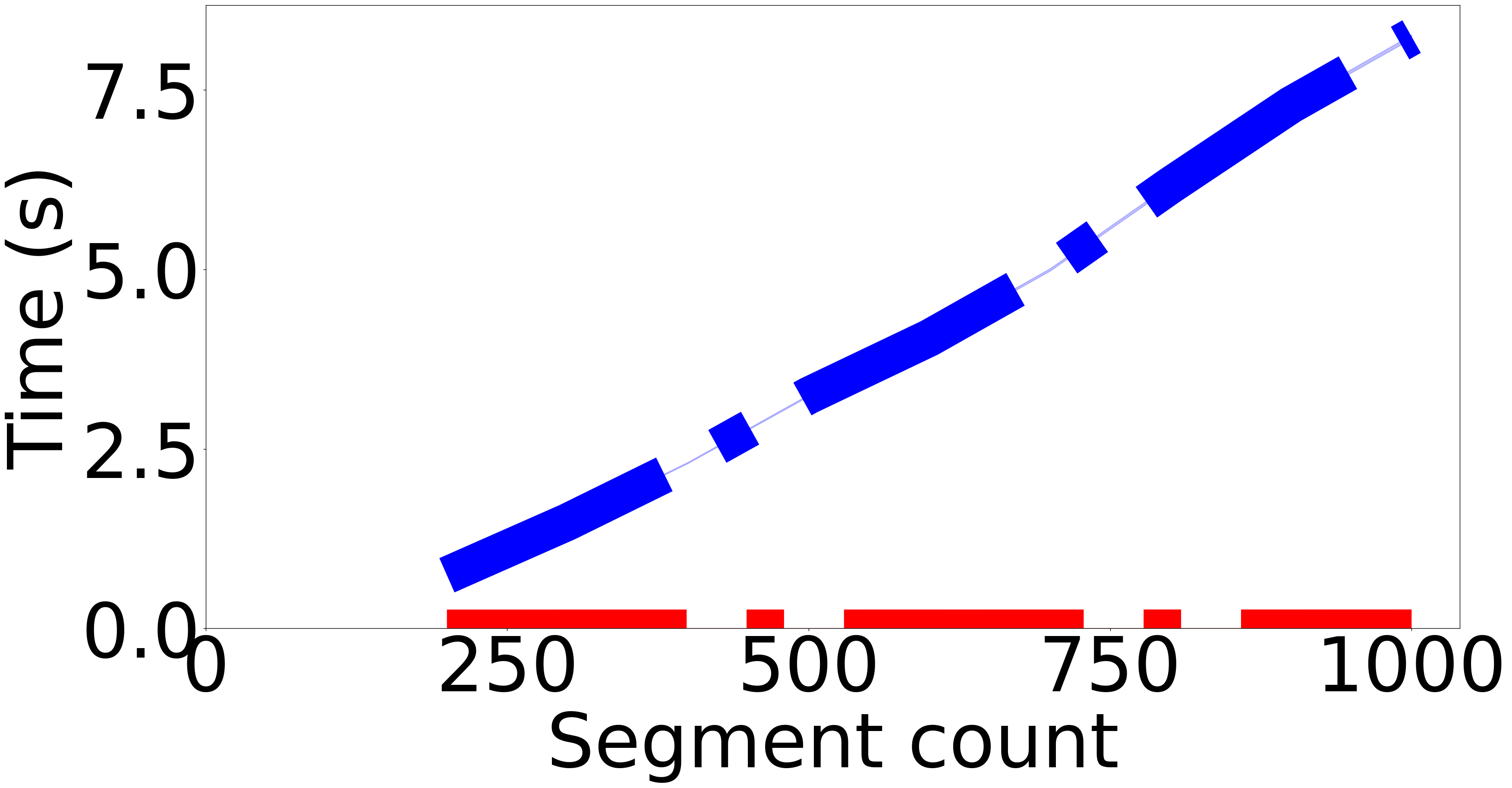}
		\includegraphics[width=0.325\textwidth]{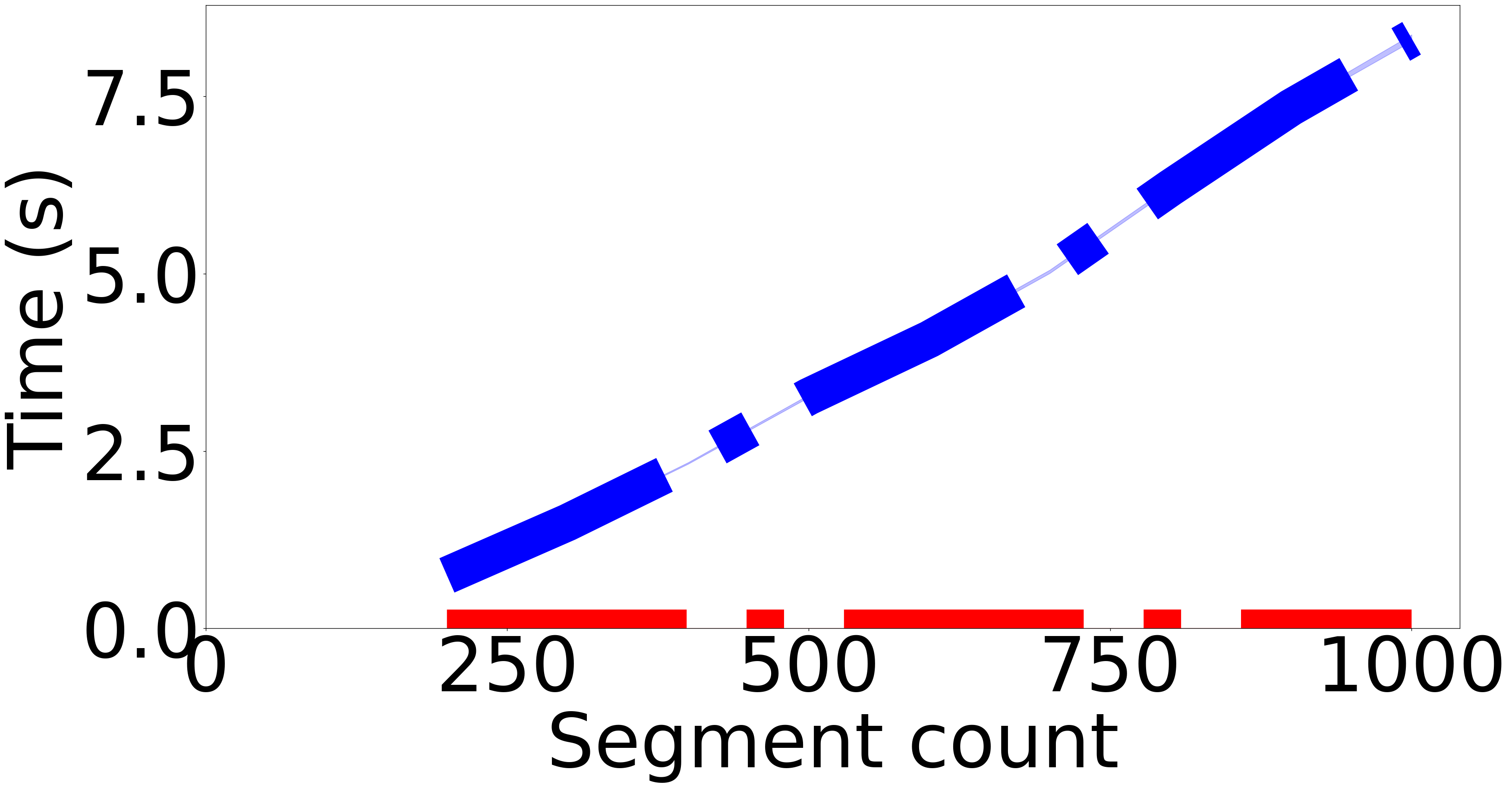}
		\caption{The results of the experiment from Section~\ref{sec:ER}.
			The top row is the legend for all the plots, the middle row corresponds to the loss, and the bottom row to the times.
			The columns correspond (left to right) to the synthetic segments (dataset \textbf{(i)}), the motion vectors (dataset \textbf{(ii)}), and the Malaysia, Singapore, and Brunei roads (dataset \textbf{(iii)}).
		}
		\label{fig: res}
	\end{figure}

	\textbf{Results.}
	Our results are presented in Figure~\ref{fig: res}.
	In all the figures, the values presented are the medians across the tests, along with error bars that present the $25\%$ and $75\%$ percentiles.
	
	As shown in Fig.~\ref{fig: res}, $\our$ consistently achieves lower loss and significantly faster runtimes compared to $\LineClust$. 
	Our runtime is dominated by the $k$-means++ call, reflecting the simplicity of our reduction.
	
	\section{Application: video tracking and detection}\label{sec: video application}
	
	\textbf{Goal.}
	These tests demonstrate that using the coreset from Algorithm~\ref{algorithm - deterministic sample} enables real-time tracking and detection of moving objects in video.
	The next paragraph provides a summary of the video tracking and detection setup.
	
	\subsection{Preliminaries on video tracking and detection}\label{sec: video Preliminaries}
	Tracking objects in RGB videos is a well-studied problem, with thousands of papers in recent years~\cite{object_detection_survey}.
	Neural networks achieve state-of-the-art results, but training and data labeling are costly, and real-time performance ($\approx 30$ fps) often requires mid-level GPUs.
	Moreover, small perturbations can ``fool’’ networks~\cite{attack_on_neural_networks}, and it is unclear whether more sophisticated attacks could bypass recent robustness improvements~\cite{Occlusion_Robust}.
	Although our method does not use object detection, real-time methods that do, e.g.,~\cite{mean-shift}, are available.
	
	Our proposed method leverages motion vectors for video tracking and detection, as detailed below.
	\subsection{Motion vectors}\label{sec: motion vectors explenation}
\begin{figure}[h]
		\centering
		\includegraphics[width=0.495\textwidth]{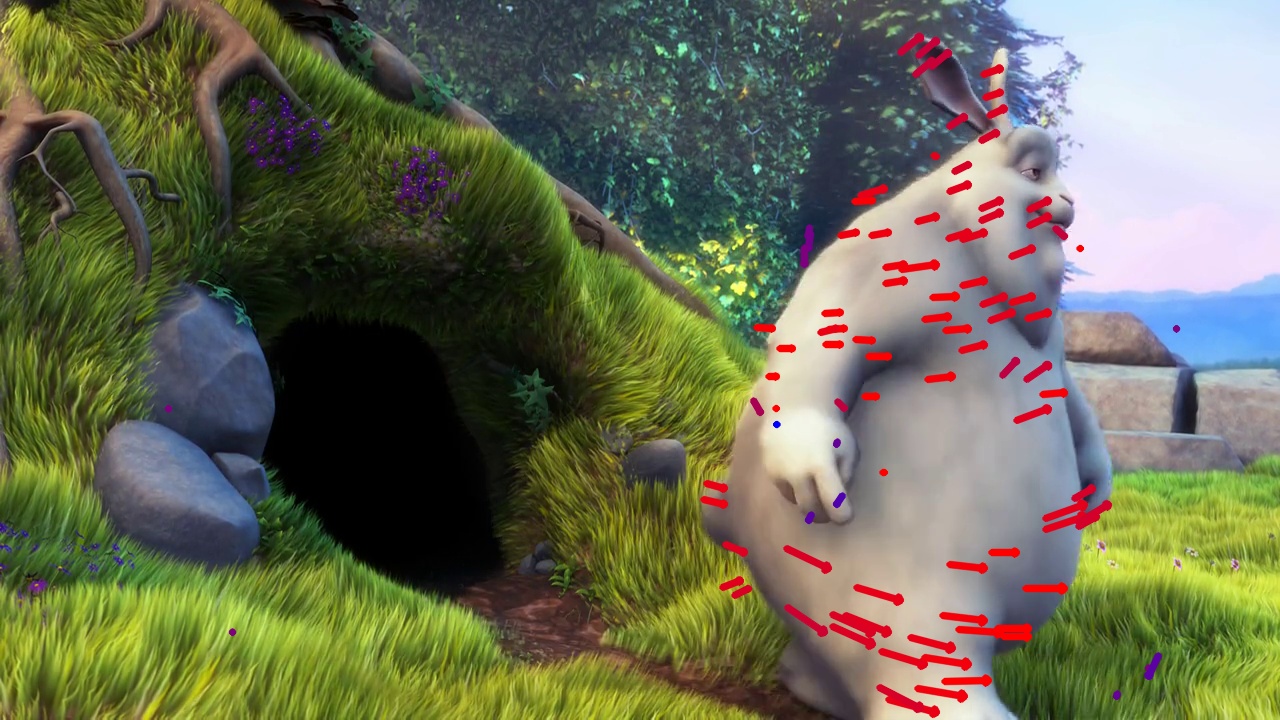}
		\includegraphics[width=0.495\textwidth]{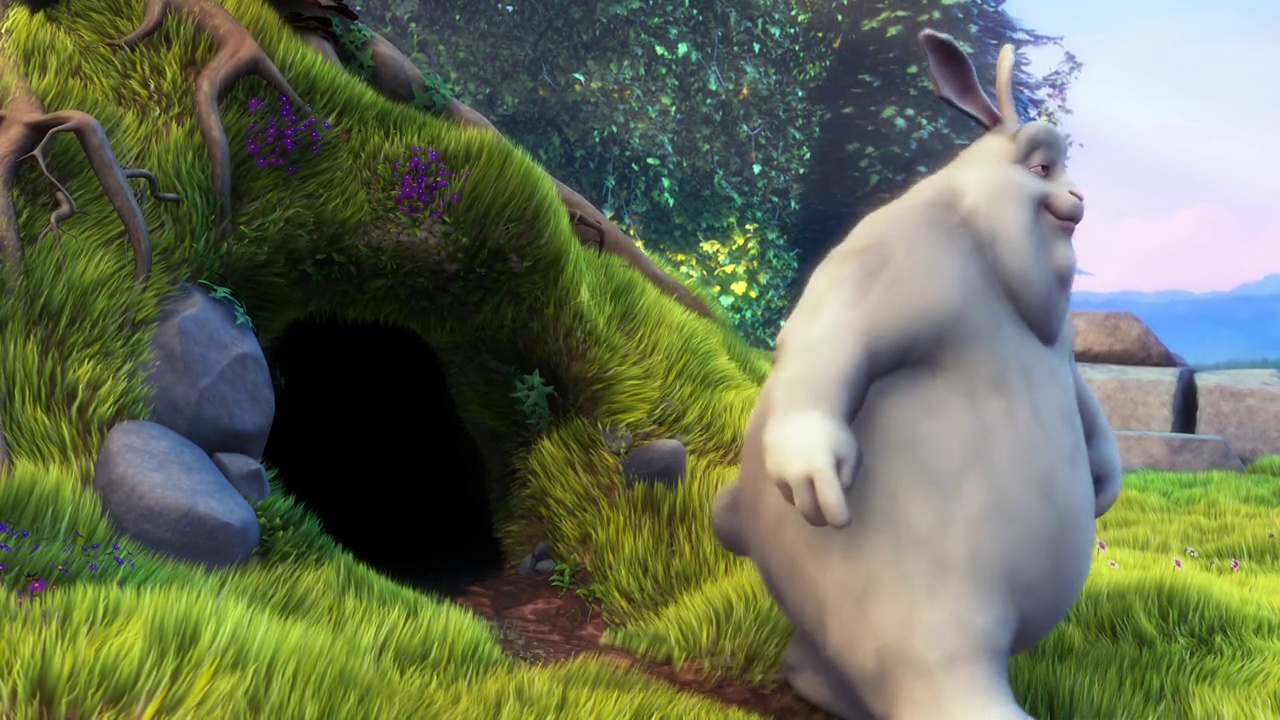}
		\caption{\textbf{Illustration of motion vectors.}
			Left: a snapshot from~\cite{Bunny_video} with a sub-sample of motion vectors to the next frame (blue: left, red: right). Right: the corresponding vectors. While most vectors follow the movement, some noise is visible, e.g., on the rock and grass in the left half.}
		\label{fig:mv_example}
	\end{figure}

	Motion vectors, computed in real time by standard video encoders such as H.264~\cite{h264} and H.265~\cite{MPEG}, map blocks from one frame to another, typically minimizing a loss function (e.g., MSE of RGB values) so that only the vectors and residual differences need to be stored. We focus on the simple case of mapping each frame to its immediate predecessor (illustrated in Figure~\ref{fig:mv_example}). Because our method utilizes only motion vectors and not RGB data, it enables real-time object tracking while preserving some degree of privacy.
	
	\subsection{The proposed method}\label{sec: proposed method}
	For simplicity, the experiments use the sum of squared distances between objects, corresponding to Gaussian noise~\cite{Gaussian_noise}.
	
	\textbf{Tracking method.}
	In general, we set $k$ equal to the number of objects tracked. For our test, tracking a single object, we use $k=2$ (for $k=1$ no clustering needs to be done and the tracking can be done as in~\cite{sumOfMotionVectors}) and track the largest cluster by the number of points assigned.
	
	For each 10 consecutive frames:\\
	\textbf{(i).} Append each motion vector’s angles to $(0,1)$ and $(1,0)$, scaled so that $180^\circ$ corresponds to the largest endpoint, yielding 4-dimensional vectors.
	\textbf{(ii).} If there are more than 1000 vectors, sample 1000 uniformly without replacement.
	\textbf{(iii).} Treat each 4-D vector as a segment and apply Algorithm~\ref{algorithm - deterministic sample} with coreset size 10.
	\textbf{(iv).} Compute $k$-means for the resulting points via~\cite{kmeans++}.
	\textbf{(v).} Track the mean start and end positions of motion vectors in the largest cluster.

	\textbf{Software.} 
	We implemented our algorithms in Python 3.8~\cite{python} utilizing \cite{opencv}, \cite{mvextract}, \cite{vidgear}, and \cite{numpy}.
	The source code is provided in~\cite{opencode}.
	
	\subsection{Big Buck Bunny test} \label{sec: bunny test}
	For this test, we used a 400-frame clip of the Big Buck Bunny video~\cite{Bunny_video} at 720×1280p, streamed in real time from a file. This video is widely used in the video tracking community and is licensed under Creative Commons Attribution 3.0, allowing free reuse with proper attribution~\cite{Bunny_video} (\url{https://peach.blender.org/about}
	). We selected this segment because it shows a bunny walking across a stationary background, allowing visual validation of tracking results.
	
	\textbf{Hardware.} 
	In this section, we used a standard Laptop with an Intel Core i3-1115G4 processor and 16GB of RAM.
	
\begin{figure}[h]
		\centering
		\includegraphics[width=0.325\textwidth]{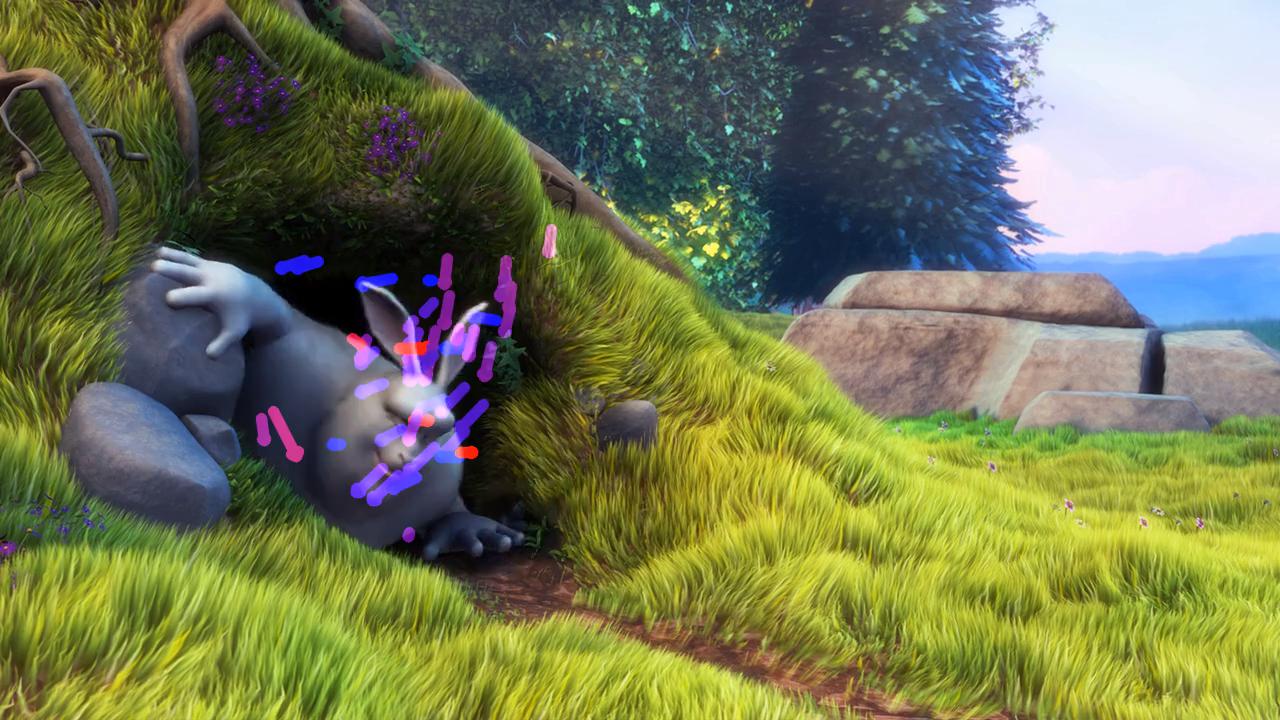}
		\includegraphics[width=0.325\textwidth]{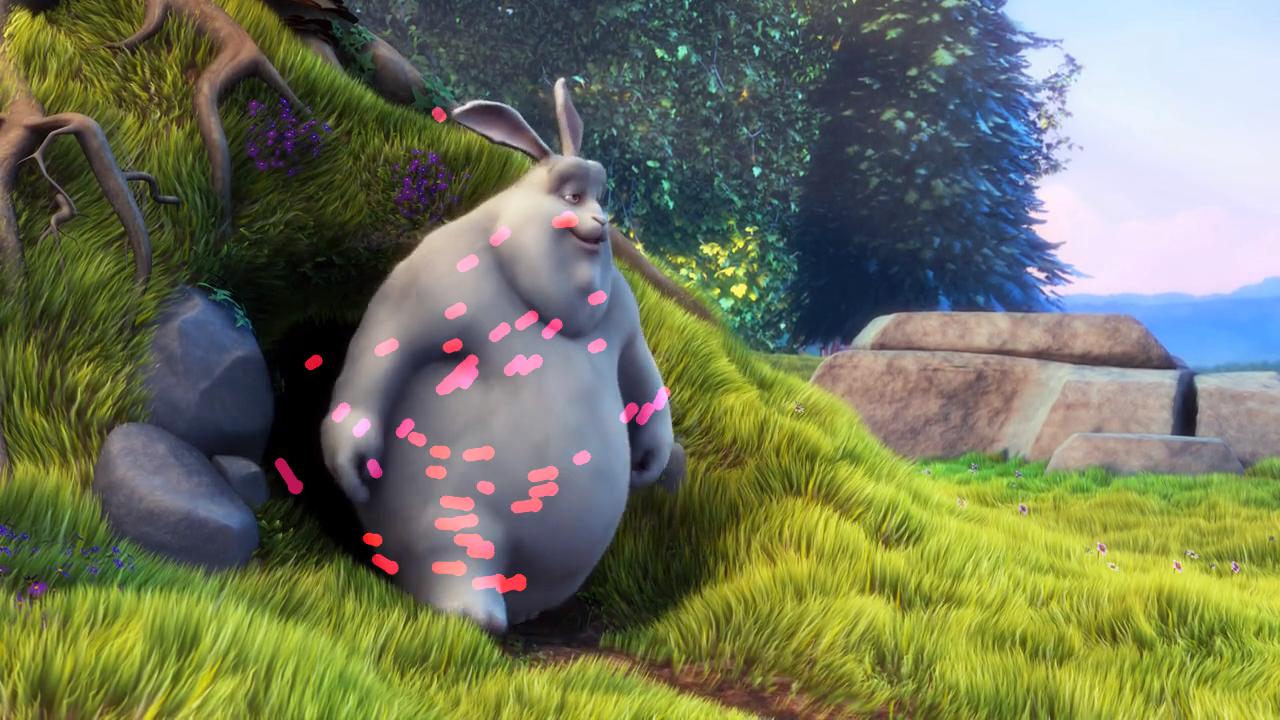}
		\includegraphics[width=0.325\textwidth]{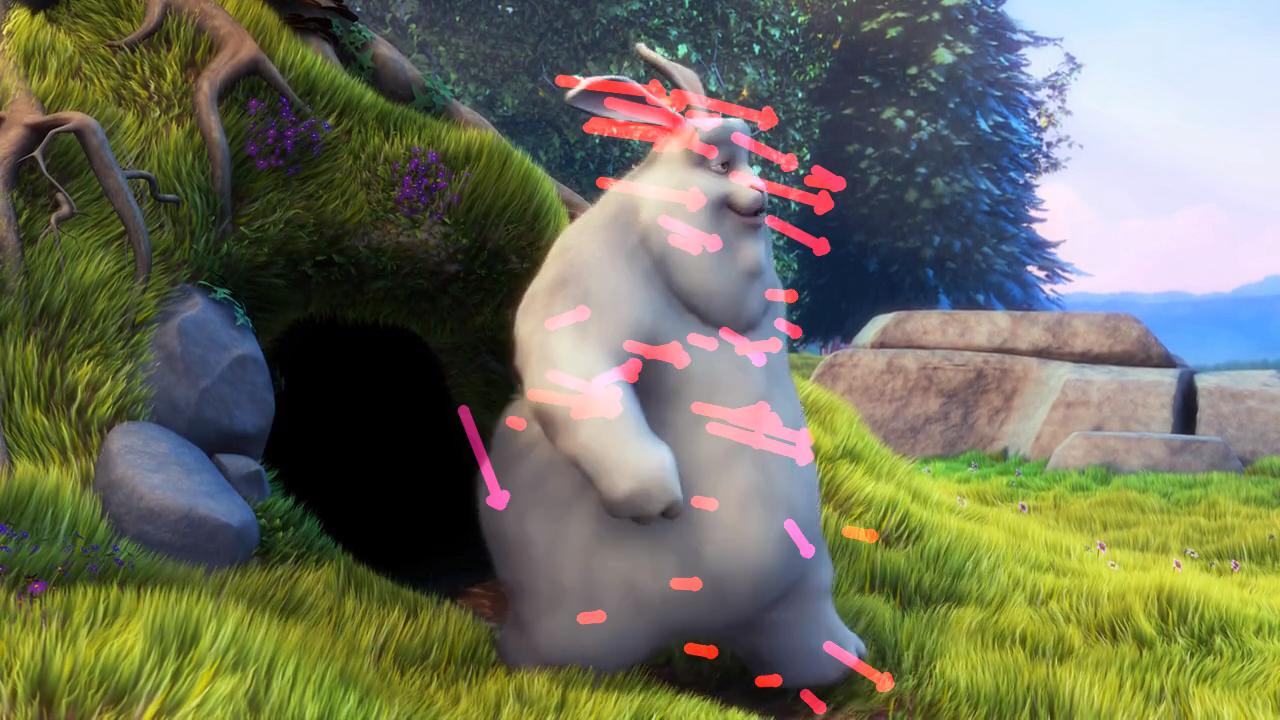}
		
		\includegraphics[width=0.325\textwidth]{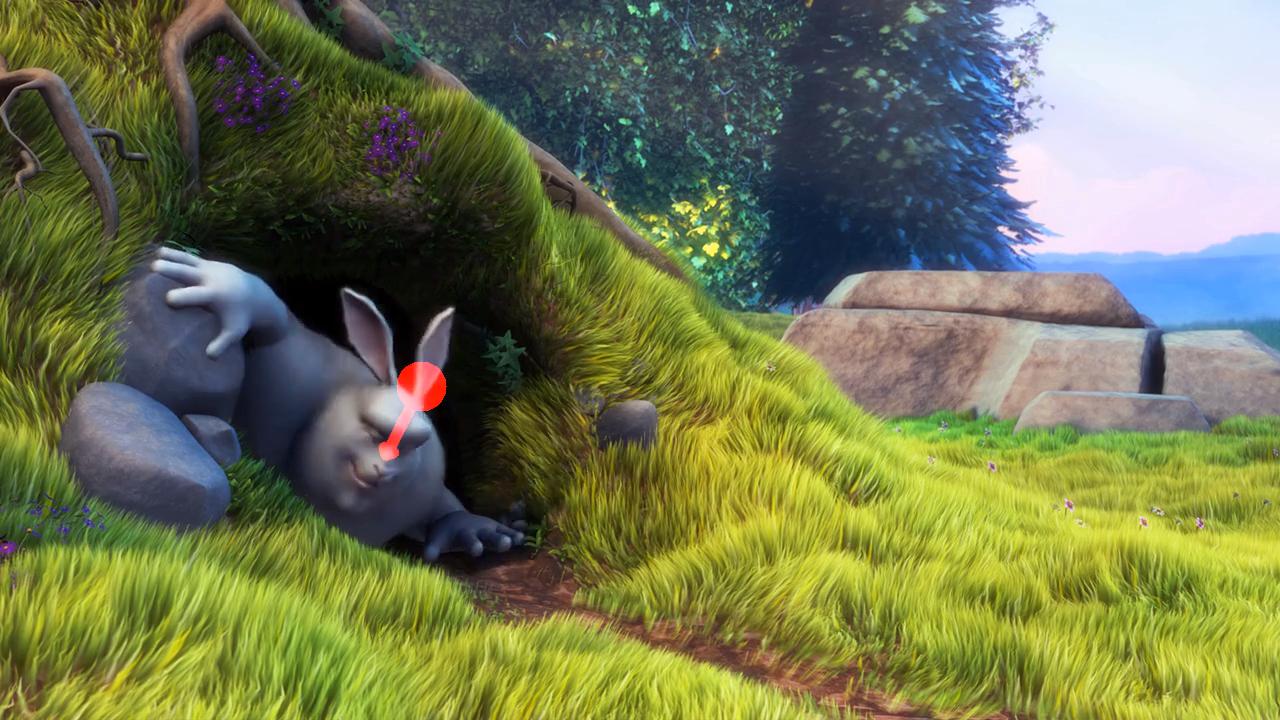}
		\includegraphics[width=0.325\textwidth]{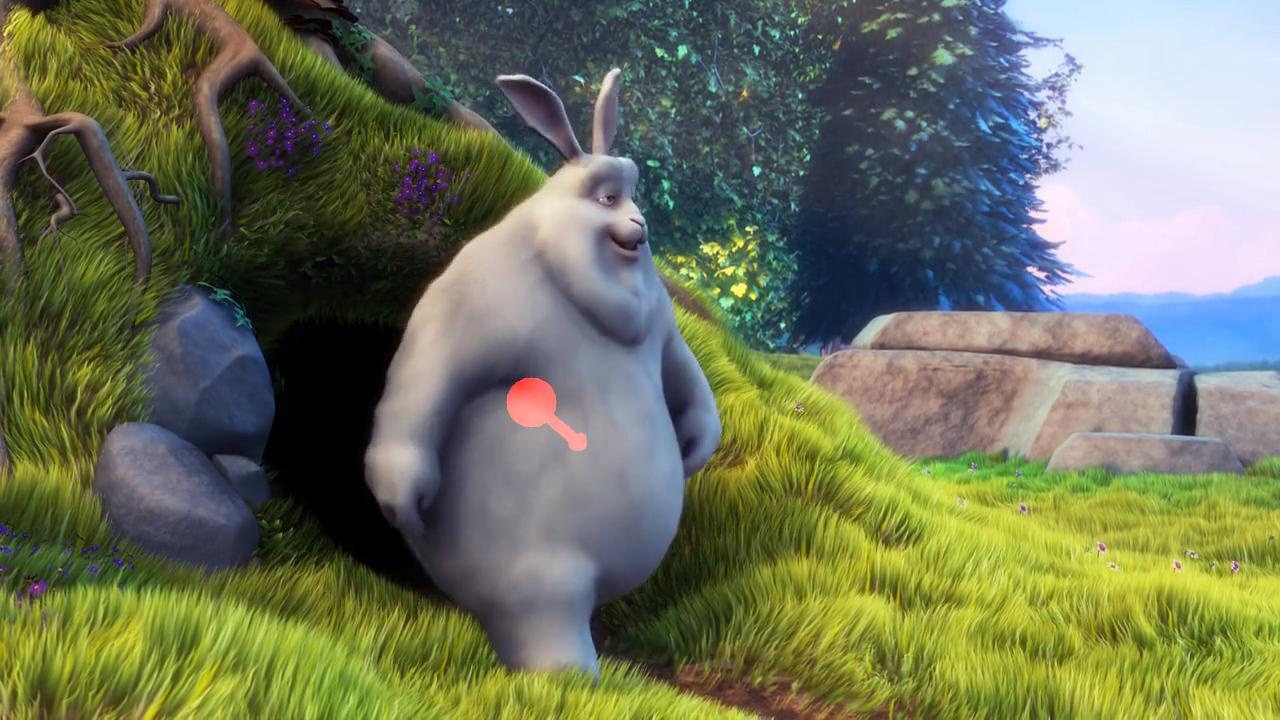}
		\includegraphics[width=0.325\textwidth]{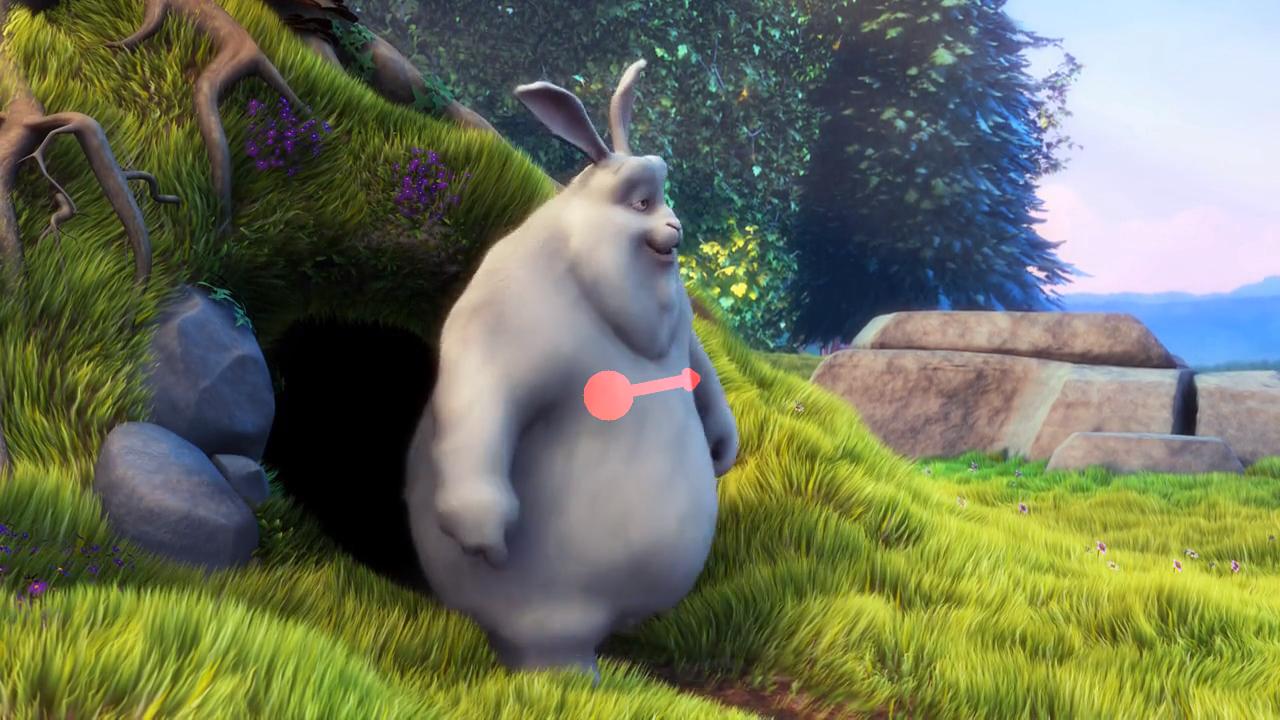}
		\caption{A subset of results from the experiment in Section~\ref{sec: bunny test}. The top row displays the motion vector clusters, and the bottom row shows the cluster centers and their corresponding mean directions. The left column captures the bunny rising from the cave, while the middle and right columns show the bunny walking right at slightly different times.}
		\label{fig: mv example}
	\end{figure}

	Figure~\ref{fig: mv example} shows examples from the clip, with the complete tracking results available in~\cite{opencode}. In the left column, the motion vectors point opposite to the bunny’s actual movement, occurring when the bunny first enters the frame, and the closest color matches are from already visible areas. Nonetheless, the cluster center still aligns with the bunny’s center, so the tracked position remains visually accurate. Once the bunny is fully in view and walking (bottom images), the predicted movement aligns with the actual motion, and the cluster center follows the object correctly.
	
	\textbf{Running time.}
	Our tracking algorithm processed the 400-frame clip in 0.28 seconds, achieving over 1,400 fps. For comparison, YOLOv8~\cite{yolov1to8} (an improvement over YOLOv5~\cite{yolov5}) takes 33.4 seconds on the same clip, or 12 fps~\cite{opencode}.

	\subsection{Re-running for Single-board computer:}\label{sec: Single-board}
	We reran the video tracking test (Section~\ref{sec: bunny test}) on the Libre Computer AML-S905X-CC (Le Potato,~\url{https://libre.computer/products/aml-s905x-cc/}), a small single-board computer similar to Raspberry Pi~\cite{raspberry}, using the official Raspberry Pi OS. Due to limited ARM support, motion vectors were pre-extracted and transferred as a Numpy array.
	This test demonstrates that our method can run in real time on low-end hardware.
	Results were essentially unchanged, with only minor differences resulting from clustering tie-breaking and sample noise. The tracking algorithm ran in 4.23 seconds for 400 frames, achieving a frame rate of over 94 fps.
	
	\subsection*{Conclusion and future work}
	This paper provides a coreset construction that gets a set of segments in $\REAL^d$ and returns a small weighted set of points (coreset) that approximates its sum of (fitting) distances to any weighted $k$-centers, up to a factor of $1\pm\eps$.
	This was achieved by reducing the problem of fitting $k$-centers to segments to the problem of fitting $k$-centers to points.
	This reduction enabled us to leverage the extensive literature on $k$-means to obtain an efficient approximation and achieve low running times in the experimental sections.
	
	While we have not analyzed the variants of the problem where an additional constraint requires each segment to be assigned to a single center, we believe that this would follow from incorporating our result with Section 15.2 of~\cite{newframework}.
	We have not considered this direction since our method is based on reducing the segments to a set of points and applying the classic $k$-means clustering, which can be computed efficiently via~\cite{kmeans++}. Even after compression, this variation on Problem~\ref{problem 1} yields a complex set clustering problem.
	
	Our results support loss functions where the integral in~\eqref{eq: problem main} at Problem~\ref{problem 1} is not necessarily elementary, and as such, direct minimization seems to us unfeasible.

	\bibliographystyle{splncs04}
	\bibliography{ref}
	\appendix
	
	\semipart{Appendix}
	
	\clearpage
	
	\section{Main characteristics of the datasets used in Section~\ref{sec:ER}}\label{sec: demo}
	Table~\ref{tab:dataset_summary} summarizes the datasets used in Section~\ref{sec:ER}.
	
	\begin{table}[h]
		\centering
		\small
		\renewcommand{\arraystretch}{1.25}
		\caption{Dataset summary.}
		\rowcolors{2}{white}{gray!15}
		\begin{tabular}{|l||c|c|c|}
			\hline
			\textbf{Dataset} & \textbf{Synthetic Segments} & \textbf{Motion Vectors} & \textbf{Road Segments} \\
			\hline
			\hline
			Source & Synthetic generation & Motion vectors from~\cite{Bunny_video} & OpenStreetMap \\
			\hline
			Size & Varies & Few seconds of video & 1M longest roads \\
			\hline
			Space & $[-1,1]^{10}$ & $0\!-\!720 \times 0\!-\!1280$ & $[-90,90]\times[-180,180]$ \\
			\hline
			Features & Float (start/end points) & Float (start/end points) & Lat/lon (deg) \\
			\hline
			Sampling & None & Weighted by length & Weighted by length \\
			\hline
			Data type & Synthetic & Animated motion & Real geospatial \\
			\hline
			Use Case & Testing & Object tracking & Road analysis \\
			\hline
			Coord. System & Euclidean & Euclidean & Geographic \\
			\hline
			License & N/A & CC BY 3.0 & ODbL \\
			\hline
		\end{tabular}
		\label{tab:dataset_summary}
	\end{table}
	
	To provide context for the datasets used in Section~\ref{sec:ER}, we present examples of the datasets along with corresponding results and additional illustrations.
	
	\begin{figure}[h!]
		\centering
		\includegraphics[width=0.495\textwidth]{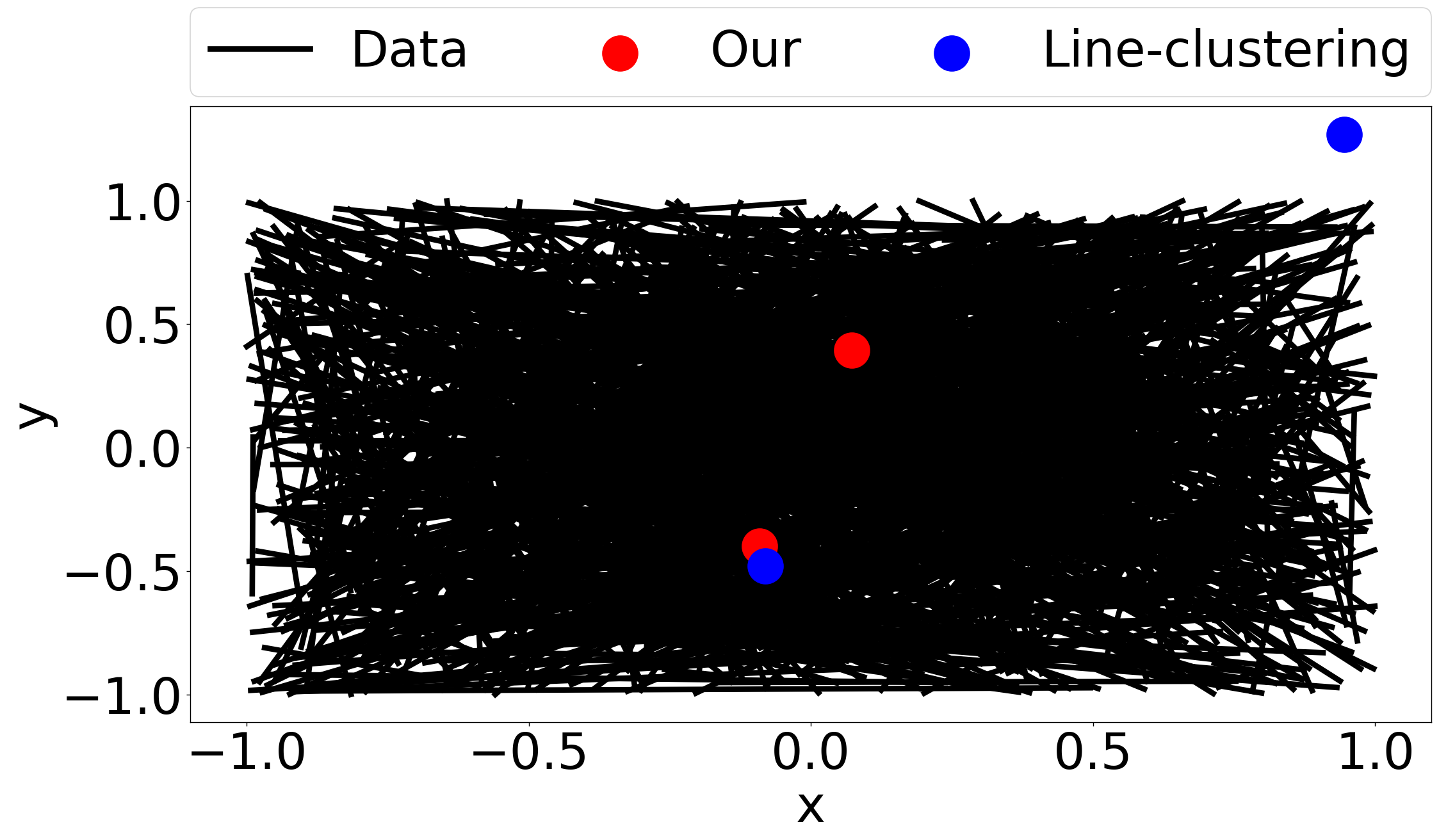}
		\includegraphics[width=0.495\textwidth]{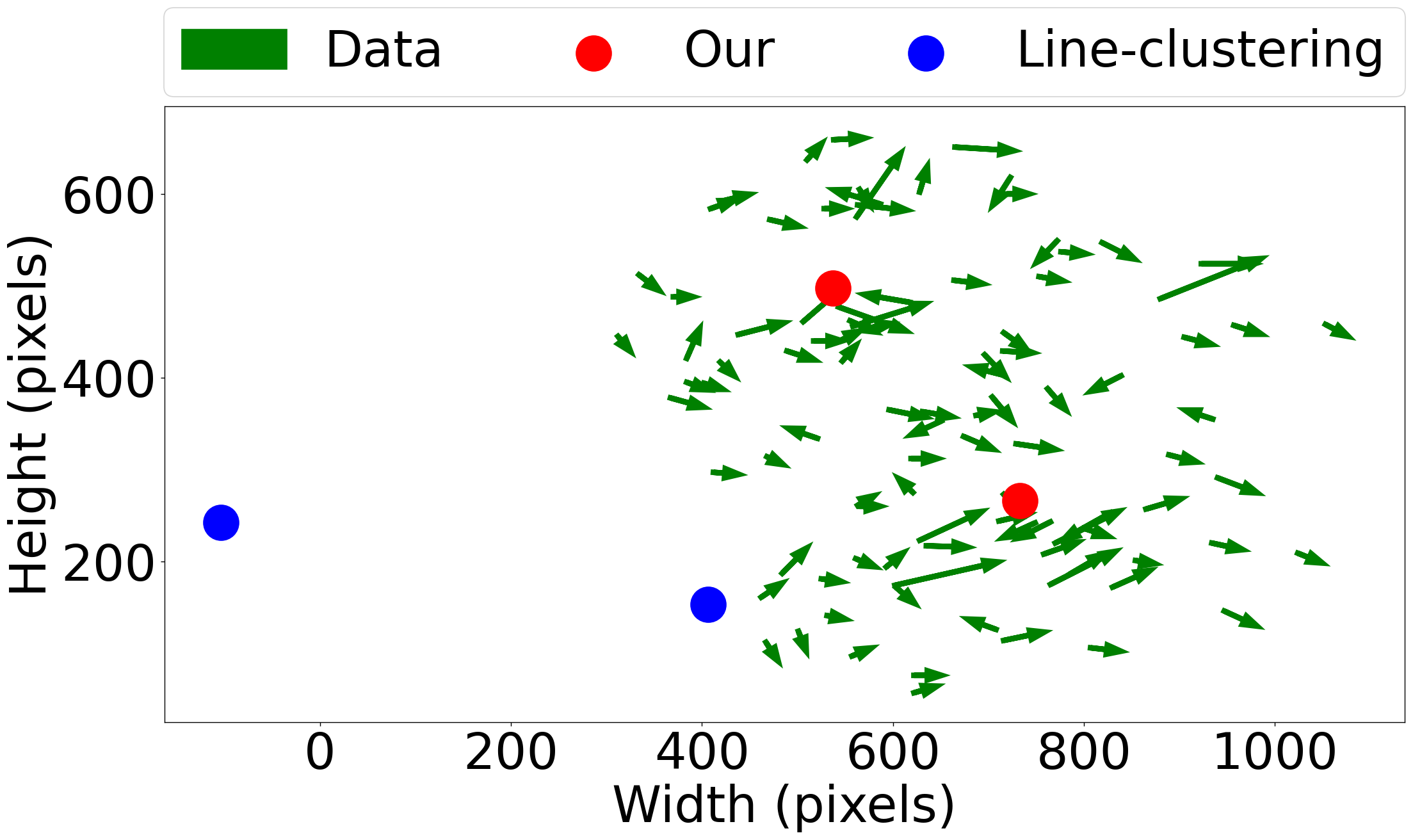}
		
		\includegraphics[width=0.495\textwidth]{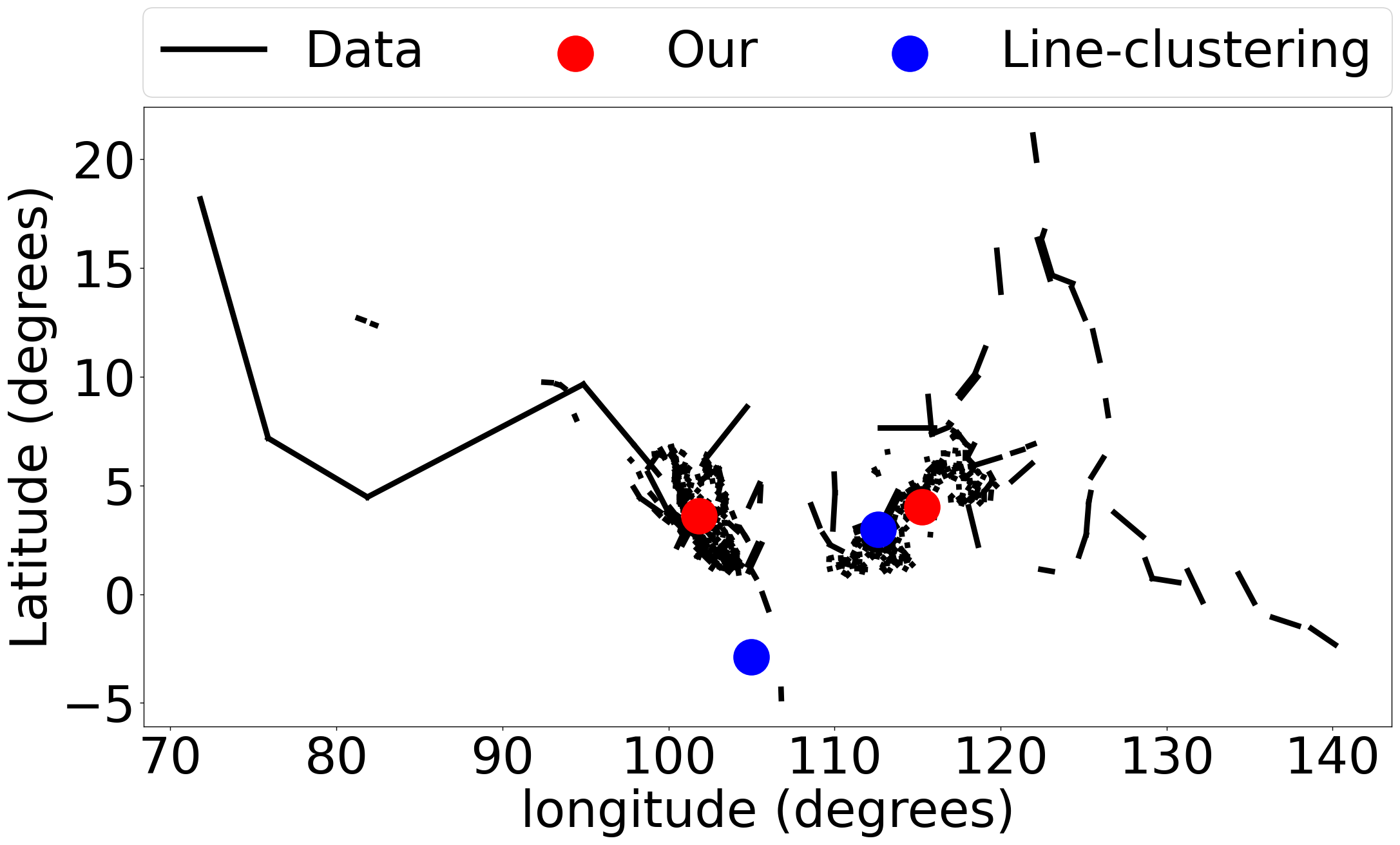}
		\includegraphics[width=0.495\textwidth]{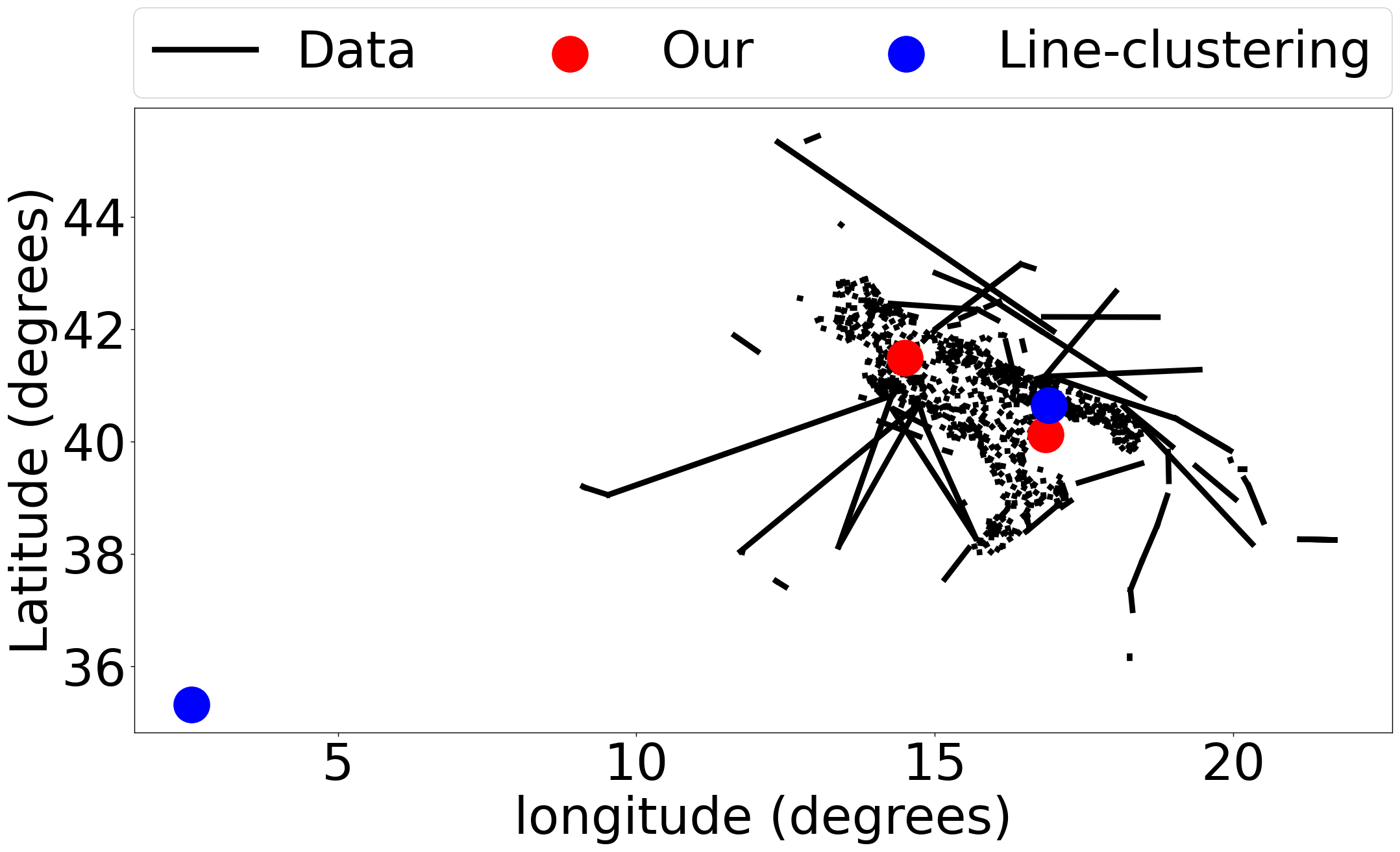}
		
		\caption{Plots of the datasets (clockwise from top left): 
			(i) synthetic segments, with the segment dimension reduced to $2$, 
			(ii) motion vectors, 
			(iii) Malaysia, Singapore, and Brunei roads, and 
			(iv) Southern Italy roads. 
			In all plots, we set the number of segments to $1000$ and show $\our$ centers (red) and $\LineClust$ centers (blue).}
		\label{fig:demo}
	\end{figure}
	
	An additional dataset is included in this demonstration:
	
	\textbf{Dataset (iv)}: An i.i.d. sample of road segments similar to dataset (iii), but restricted to southern Italy. 
	We selected this region because of its distinctive ``boot" shape, which allows for a quick visual check of the road layout without requiring extensive geographical knowledge or consulting maps.
	
	Figure~\ref{fig:demo} confirms the motivations for the chosen datasets. 
	As expected, the $\LineClust$ centers are often far from the segments, since this method extends each segment to an infinite line, whereas $\our$ better captures the actual segment locations.
	
	\section{Bounding the sensitivity of one $r$-Lipschitz function:}\label{sec: sen bound}
	In this section, we bound the sensitivity of a single $r$-Lipschitz function.
	In the following sections, we will extend this to a sensitivity bound for the minimum over $r$-symmetric functions.
	To simplify the following lemma, we prove the following propositions.
	\begin{proposition} \label{power sum proposition}
		For every $r \geq 0$, and an integer $n\geq 2$ we have 
		\[
		\sum_{i =1}^n i^{r} \geq (n/2)^{r+1}.
		\]
	\end{proposition}
	\begin{proof}
		Let $r\geq 0$, and let $n\geq 2$ be an integer.
		We have
		\begin{equation*}
			\sum_{i= 1}^n i^{r} \geq \sum_{i= \lceil n/2\rceil}^n i^{r}\geq  \lceil n/2\rceil \cdot \big(n-\lfloor n/2\rfloor\big)^{r} = \lceil n/2\rceil^{r+1} \geq (n/2)^{r+1},
		\end{equation*}
		where the second inequality is by observing that the summation is over $\br{\lceil n/2\rceil,\cdots,n}$.
	\end{proof}
	
	\begin{proposition} \label{log other side inequality proposition}
		Let $r\geq 0$, and $f$ be an $r$-log-Lipschitz function; see Definition~\ref{def: log-Lipschitz}.
		For every $x,x' >0$ we have
		\begin{equation} \label{sen bound eq helper 1}
			f(x) \cdot \min \left\{1,\left(\frac{x'}{x}\right)^{r}\right\}\leq  f(x').
		\end{equation}
	\end{proposition}
	\begin{proof}
		If $(x/x')\geq 1$ we have
		\begin{equation*}
			f(x) \cdot \left(\frac{x'}{x}\right)^{r} \leq  f(x'),
		\end{equation*}
		which follows from substituting $c:=(x/x'),x:=x',h:=f$ in the definition of $r$-log-Lipschitz functions, and dividing both sides by $(x/x')^{r}$.
		If $(x/x')\leq 1$, we have $x\leq x'$.
		By the definition of $r$-log-Lipschitz functions, we have that $f$ is non-decreasing.
		Hence, $f(x)\leq f(x')$.
		Therefore \eqref{sen bound eq helper 1} holds.
	\end{proof}
	
	\begin{lemma} \label{mini sen bound}
		Let $n\geq 2$ be an integer.
		Let $r\geq 0$, and let $F$ be the set of all the $r$-log-Lipschitz functions; see Definition~\ref{def: log-Lipschitz}.
		For every $x\in \br{1,\cdots,n}$ we have that
		\begin{equation*}
			\max_{f \in F} \frac{f(x)}{\displaystyle \sum_{x' =1}^n f(x')} \leq \frac{2^{r+2}}{n}.
		\end{equation*}
	\end{lemma}
	\begin{proof}
		Let $f\in F$.
		By Proposition~\ref{log other side inequality proposition}, for every  $x,x' \in \br{1,\cdots,n}$ we have
		\begin{equation}\label{sen bound eq 1}
			f(x) \cdot \min \left\{1,\left(\frac{x'}{x}\right)^{r}\right\} \leq  f(x').
		\end{equation}
		
		Hence, for every $x \in \br{1,\cdots,n}$ and any $f\in F$ we have
		\begin{align}
			\label{sen bound eq 2.1}
			\frac{f(x)}{\displaystyle \sum_{x' =1}^n f(x')} & \leq \frac{f(x)}{\displaystyle \sum_{x' =1}^n \left(f(x)\cdot \min\left\{1,\left(\frac{x'}{x}\right)^{r}\right\} \right)} \\
			\label{sen bound eq 2.2}
			& = \frac{1}{\displaystyle \sum_{x' =1}^n \min\left\{1,\left(\frac{x'}{x}\right)^{r}\right\}} \\
			\label{sen bound eq 2.3}
			& = \frac{1}{\displaystyle \sum_{x' = x+1}^n (1) + \sum_{x'=1}^x \left(\frac{x'}{x}\right)^{r}} \\
			\label{sen bound eq 2.4}
			& \leq \min \left\{ \frac{1}{\displaystyle \sum_{x'  = x+1}^n (1)} , \frac{x^r}{\displaystyle \sum_{x'=1}^x \left(x'^{r}\right)} \right\}.
		\end{align}
		where \eqref{sen bound eq 2.1} is by plugging \eqref{sen bound eq 1}, \eqref{sen bound eq 2.2} is by dividing both the numerator and denominator by $f(x)$, \eqref{sen bound eq 2.3} and \eqref{sen bound eq 2.4} are by reorganizing the expression.
		If $x \leq n/2$ by \eqref{sen bound eq 2.1} to \eqref{sen bound eq 2.4} we have
		\begin{equation} \label{sen bound eq 3}
			\frac{f(x)}{\displaystyle \sum_{x' =1}^n f(x')} \leq \frac{1}{\displaystyle \sum_{x'=x+1}^n 1} = \frac{1}{|n-x-1|} \leq \frac{2}{n}.
		\end{equation}
		If $x\geq n/2$, by \eqref{sen bound eq 2.1} to \eqref{sen bound eq 2.4} we have
		\begin{align}
			\label{sen bound eq 4.1}
			\frac{f(x)}{\displaystyle \sum_{x' =1}^n f(x')} & \leq \frac{x^r}{\displaystyle \sum_{x'=1}^x \left(x'^{r}\right)}\\
			\label{sen bound eq 4.2}
			& \leq \frac{2^{r+1} x^r}{x^{r+1}}\\
			\label{sen bound eq 4.3}
			& = \frac{2^{r+1}}{x} \\
			\label{sen bound eq 4.4}
			& \leq \frac{2^{r+2}}{n},
		\end{align} 
		where \eqref{sen bound eq 4.1} is by \eqref{sen bound eq 2.1} to \eqref{sen bound eq 2.4}, \eqref{sen bound eq 4.2} is since $x \geq n/2$ by the definition of the case and plugging $n=n/2$ in Proposition~\ref{power sum proposition}, \eqref{sen bound eq 4.3} is by reorganizing the expression, \eqref{sen bound eq 4.4} is since $x\geq n/2$ (the definition of the case).
		Combining \eqref{sen bound eq 3} and \eqref{sen bound eq 4.1} to \eqref{sen bound eq 4.4} proves the lemma.
	\end{proof}
	
	The main idea in the previous lemma can be demonstrated using the following Figures.
	
	\begin{figure}[h]
		\centering
		\includegraphics[width=0.3275\textwidth]{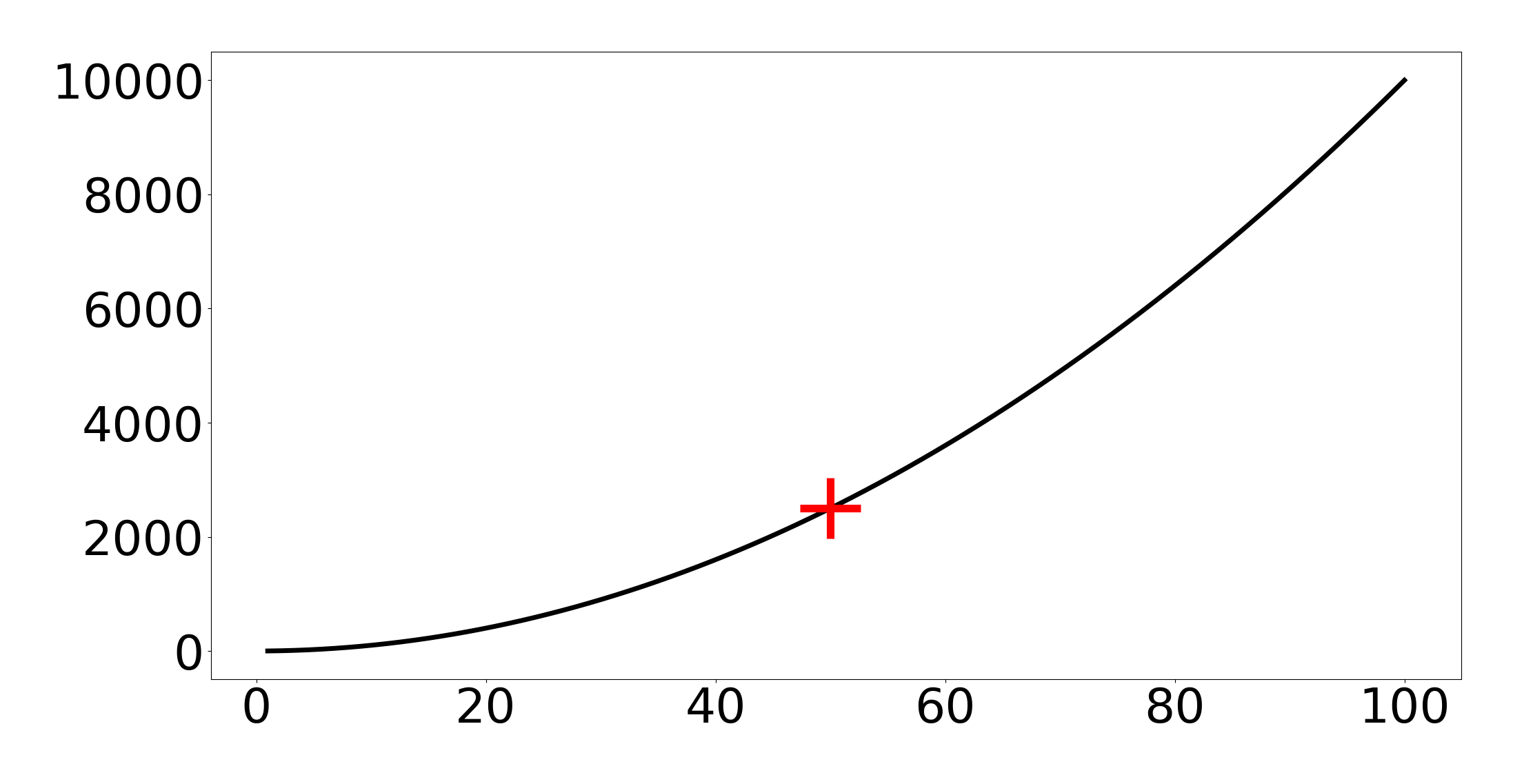}
		\includegraphics[width=0.3275\textwidth]{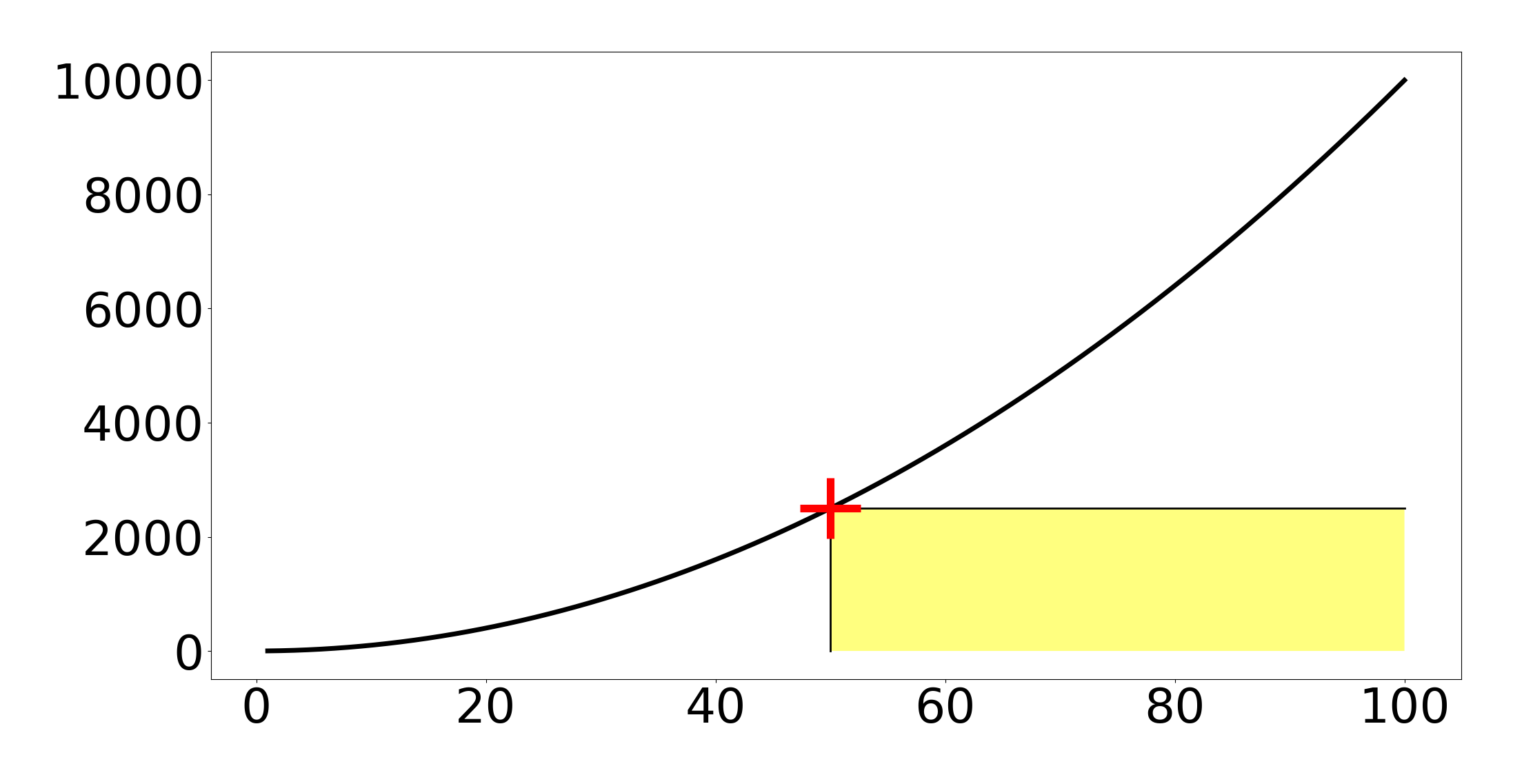}
		\includegraphics[width=0.3275\textwidth]{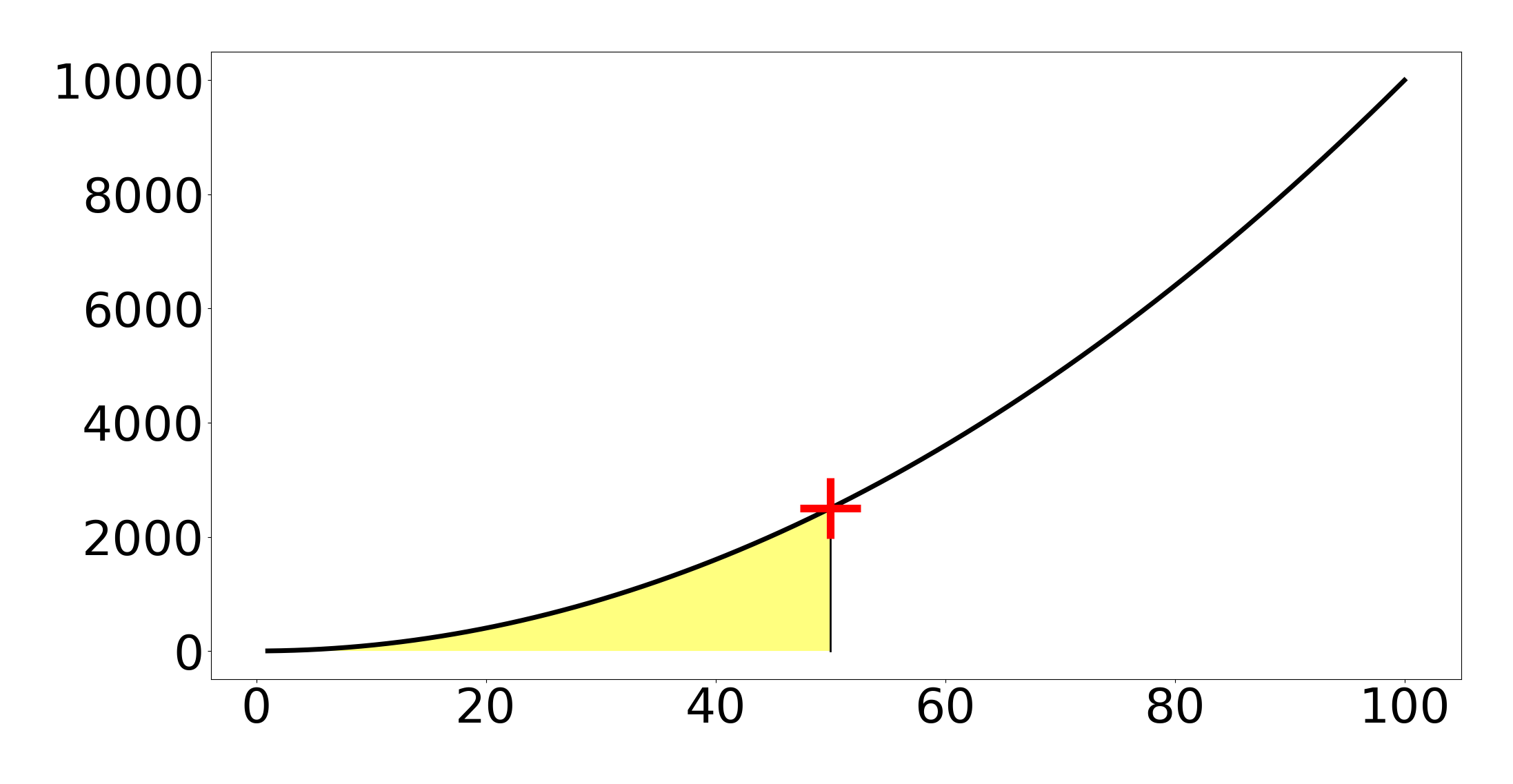}
		\caption{
			\textbf{Left.} Illustration of a point (red) $p=(50,2500)$ on the plot of the $2$-log-Lipschitz function (black, see Definition~\ref{def: log-Lipschitz}) $f:[0,\infty) \to \REAL$, which maps every $x\in [0,\infty)$ to $f(x) = x^2$.
			\textbf{Middle.} Demonstration of the bound for the points with $x$-value at most $50$, which follows by charging against $f(x)$ utilizing that $r$-log-Lipschitz functions are non-decreasing .
			\textbf{Right.} Demonstration of the bound for the points with $x$-value larger than $50$, bounded using Proposition~\ref{power sum proposition} due to $f$ being $2$-log-Lipschitz.}
	\end{figure}
	
	\section{Sensitivity bound for the minimum over symmetric-$r$ functions}\label{sec: sen minimum over}
	In this section, we utilize methods from the previous section to bound the sensitivity of the minimum over symmetric-$r$ functions; see Definition~\ref{def: symmetric-$r$ function}. 
	As often in prior work, e.g.~\cite{outliers-resistance}, the coreset size generally depends on the number of functions in the minimum. However, for evenly spaced data, which was not the case in~\cite{outliers-resistance}, we obtain a significantly smaller and simpler coreset.
	
	To simplify the following lemma, we will prove the following proposition.
	\begin{proposition} \label{reduce sen bound helper}
		Let $r,a\geq 0$, and let $f: [0,\infty) \to [0,\infty)$ be an $r$-log-Lipschitz function; see Definition~\ref{def: log-Lipschitz}.
		Let $\Tilde{f}: [0,\infty) \to [0,\infty)$ s.t. for every $x\geq 0$ we have $\Tilde{f}(x) := f(x+a)$, for $a> 0$.
		We have that $\Tilde{f}$ is an $r$-log-Lipschitz function.
	\end{proposition}
	\begin{proof}
		For every $x\geq 0$ and any $c\geq 1$ we have that
		\begin{equation*}
			\Tilde{f}(x\cdot c) = f(x\cdot c +a) \leq f\big( (x+a)\cdot c \big) \leq c^{r} \cdot f(x+a) = c^{r} \cdot \Tilde{f}(x),
		\end{equation*}
		where the first equality is by the definition of $\Tilde{f}$, the first inequality is since $f$ is non-decreasing, and since $a\ge 0$ and $c\ge 1$, we have $a \le a\cdot c$, the second inequality is by the definition of $f$ as an $r$-log-Lipschitz function, and the second equality is by the definition of $\Tilde{f}$. 
		Since $f$ is non-decreasing, we have that $\Tilde{f}$ is non-decreasing, which, combined with the previous result, proves the proposition.
	\end{proof}
	For simpler use of the results for the problem in Definition~\ref{def: loss function}, we prove the following lemma before the general sensitivity bound.
	\begin{lemma} \label{general sen bound part 1}
		Let $n,k\geq 1$ be integers, where $n\geq 10 k$, and let $r\geq 0$.
		Let $\Tilde{f}$ be a symmetric-$r$ function; see Definition~\ref{def: symmetric-$r$ function}.
		For every set $X\subset \br{1,\cdots,n}$ of size at least $n/k$ and any value $\Tilde{x}\in \br{1,\cdots,n}$ we have 
		\begin{equation} \label{weak reduce sen th eq:1}
			\frac{\Tilde{f}(\Tilde{x})}{\displaystyle \sum_{x'\in X} \Tilde{f}(x')} \leq \frac{(10k)^{r+1}}{n}. 
		\end{equation}
	\end{lemma}
	\begin{proof}
		
		Since $\Tilde{f}$ is a symmetric-$r$ function, there is $a \in \REAL$ and $f$ that is an $r$-log-Lipschitz function, such that for every $x\in \br{1,\cdots,n}$ we have $\Tilde{f}(x) = f\big( |x-a|\big)$.
		Let $\Tilde{x}\in \br{1,\cdots,n}$ and $x=|\Tilde{x}-a|$.
		We have that
		\begin{equation} \label{weak reduce sen th eq:2.1}
			\Tilde{f}(\Tilde{x}) = f\big( |\Tilde{x}-a|\big) = f(x).
		\end{equation}
		Let $X_1 = \br{\big\lfloor|x-a|\big\rfloor \mid x\in X,x\leq a},X_2 = \br{\big\lfloor|x-a|\big\rfloor \mid x\in X,x\geq a}$, and $X_i$ be the largest set among $X_1\setminus\br{0}$ and $X_2\setminus\br{0}$.
		We have
		\begin{align}
			\label{weak reduce sen th eq:2.2.1}
			\sum_{x'\in X} \Tilde{f}(x') &= \sum_{x'\in X} f\big(|x'-a|\big)\\
			\label{weak reduce sen th eq:2.2.2}
			&\geq \sum_{x'\in X_i} f(x'),    
		\end{align}
		where \eqref{weak reduce sen th eq:2.2.1} is since for every $x'\in \br{1,\cdots,n}$ we have that $\Tilde{f}(x) = f\big(|x'-a|\big)$ (from the definition of $f$ and $a$), \eqref{weak reduce sen th eq:2.2.2} is since $f$ is non-decreasing and $X_i \subset \br{\big\lfloor|x-a|\big\rfloor \mid x\in \Tilde{X}_i}$.
		Combining \eqref{weak reduce sen th eq:2.1} with \eqref{weak reduce sen th eq:2.2.1} and~\eqref{weak reduce sen th eq:2.2.2} yields
		\begin{equation} \label{weak reduce sen th eq:2}
			\frac{\Tilde{f}(\Tilde{x})}{\displaystyle \sum_{x'\in X} \Tilde{f}(x')} \leq \frac{f(x)}{\displaystyle \sum_{x'\in X_i} f(x')}.
		\end{equation}
		In the following, we will show that we can consider only the case where $a\in [1,n]$.
		
		If $a >n$, for every $x'\in [1,n]$, we have $|x'-a| = |(x'-n)+(n-a)| = |x'-n|+|n-a|$, follows since $x'-n,n-a \leq 0$.
		If $a <1$, for every $x'\in [1,n]$, we have $|x'-a| = |(x'-1)+(1-a)| = |x'-1|+|1-a|$, follows since $x'-1,1-a \geq 0$.
		Hence, if $a\not\in (1,n)$ there are $a'\in \br{1,\cdots,n},b'>0$ such that for every $x'\in \br{1,\cdots,n}$ we have $|x'-a| = |x'-a'|+b'$.
		Therefore, by Proposition~\ref{reduce sen bound helper} (with substitutions $\tilde{f}:=f$, $f:=f'$, $a:=b'$, and later $x:=|x'-a|$), there exists an $r$-log-Lipschitz function $f'$ and $a' \in [1,n]$ such that for every $x' \in \br{1,\cdots,n}$, we have $f(|x'-a|) = f'(|x'-a'|)$. Hence, we can assume without loss of generality that $a \in [1,n]$.

		Since $|X| \geq \lceil n /k\rceil,X\subset \br{1,\cdots,n},a \in\br{1,\cdots,n} $, by the choice of $X_i$ we have that $X_i\subset \br{1,\cdots,n}$ and $|X_i|\geq \lceil n /(2k) -1 \rceil \geq \lceil n /(2.5k) \rceil$ (using the assumption that $n\geq 10k$).
		For every $x'\in X_i$, by assigning $x',x$ and $f$ in Proposition~\ref{log other side inequality proposition}, we have 
		\begin{equation} \label{weak reduce sen th eq:3}
			f(x) \cdot \min \big\{1,(x'/x)^{r}\big\}\leq  f(x').
		\end{equation}
		Let $X' = \br{x'\in X_i \mid x'\geq x}$.
		We have %(if $X_i =X'$ jump \eqref{weak reduce sen th eq:4.2} expression to \eqref{weak reduce sen th eq:5})
		\begin{align}
			\label{weak reduce sen th eq:4.1}
			\frac{f(x)}{\displaystyle \sum_{x'\in X_i} f(x')} &\leq \frac{f(x)}{\displaystyle \sum_{x'\in X_i} \Big(f(x)\cdot \min\big\{1,(x'/x)^{r}\big\} \Big)} \\
			\label{weak reduce sen th eq:4.2}
			& = \frac{1}{\displaystyle \sum_{x'\in X_i} \min\big\{1,(x'/x)^{r}\big\}},
		\end{align}
		where \eqref{weak reduce sen th eq:4.1} is by \eqref{weak reduce sen th eq:3}, and \eqref{weak reduce sen th eq:4.2} is by dividing both sides by $f(x)$.\\
		\textbf{Case 1, if $|X'| \geq |X_i|/2$.}
		By \eqref{weak reduce sen th eq:4.1},\eqref{weak reduce sen th eq:4.2}, and that $|X_i|\geq n/(2.5k)$ we have
		\begin{equation} \label{weak reduce sen th eq:5}
			\frac{f(x)}{\displaystyle \sum_{x'\in X_i} f(x')} \leq \frac{1}{|X'|} \leq \frac{2}{|X_i|} \leq \frac{5 k}{n}.
		\end{equation}
		\textbf{Case 2, if $|X'| <|X_i|/2$.}
		By reorganizing \eqref{weak reduce sen th eq:4.2} we have
		\begin{align}
			\label{weak reduce sen th eq:4.3}
			\frac{1}{\displaystyle \sum_{x'\in X_i} \min\big\{1,(x'/x)^{r}\big\}} & = \frac{1}{\displaystyle |X'| + \frac{1}{x^r} \cdot \sum_{x'\in X_i \setminus X'} \left(x'\right)^r} \\
			\label{weak reduce sen th eq:4.4}
			& \leq \frac{x^r}{\displaystyle \sum_{x'\in X_i \setminus X'} \left(x'\right)^r }.
		\end{align}
		We have $|X_i\setminus X'| \geq |X_i|/2 \geq n/(5k)$; since $|X'| \leq |X_i|/2$ and $|X_i|\geq n/(2.5k)$.
		Since $(a,\Tilde{x})\in [1,n]^2$, we have $x=|\Tilde{x}-a| \leq n$; by the definitions of $a,\Tilde{x}$, and $x$.
		
		Then, we have
		\begin{align}
			\label{weak reduce sen th eq:6.1}
			\frac{f(x)}{\displaystyle \sum_{x'\in X_i} f(x')} & \leq \frac{x^r}{\displaystyle \sum_{x' \in X_i \setminus X'} \left(x'\right)^r} \\
			\label{weak reduce sen th eq:6.2}
			& \leq 
			\frac{x^r}{\displaystyle \sum_{x' =1}^{\lceil n/(5k)\rceil} \left(x'\right)^r}\\
			\label{weak reduce sen th eq:6.3}
			&\leq \frac{x^r}{\big(n/(10k)\big)^{r+1}}\\
			\label{weak reduce sen th eq:6.4}
			&\leq \frac{(10k)^{r+1}}{n},
		\end{align}
		where \eqref{weak reduce sen th eq:6.1} is by (\ref{weak reduce sen th eq:4.1}--\ref{weak reduce sen th eq:4.2}) and (\ref{weak reduce sen th eq:4.3}--\ref{weak reduce sen th eq:4.4}), \eqref{weak reduce sen th eq:6.2} is since $|X_i\setminus X'| \geq n/(5k),(X_i \setminus X') \subset [n]$ and $f'(x') = \left(x'\right)^r$ is an increasing function for $x\geq 0$, \eqref{weak reduce sen th eq:6.3} is from plugging $n=n/(5k)$ in Proposition~\ref{power sum proposition}, and \eqref{weak reduce sen th eq:6.4} is by assigning that $x\leq n$ and rearranging.
		
		Combining \eqref{weak reduce sen th eq:2}, \eqref{weak reduce sen th eq:5}, and (\ref{weak reduce sen th eq:6.1}--\ref{weak reduce sen th eq:6.4}) proves the lemma.
	\end{proof}
	In the following lemma, we provide the desired sensitivity bound.
	%, which plugging in Theorem~\ref{theorem - theorem-the output of coreset is coreset} yields the general coreset construction, which is essentially a sufficiently large uniform sample.
	\begin{lemma}\label{general sen bound}
		Let $n,k\geq 1$ be integers, such that $n\geq 10 k$, and let $r\geq 0$.
		Let $F\subset \br{f:\br{1,\cdots,n} \to (0,\infty)}$, be a set of $|F|=k$ symmetric-$r$ functions; see Definition~\ref{def: symmetric-$r$ function}.
		For every $x\in \br{1,\cdots,n}$, let $\displaystyle f(x) := \min_{f'\in F} f'(x)$.
		Then, for every $x\in \br{1,\cdots,n}$ we have 
		\begin{equation*}
			\frac{f(x)}{\displaystyle \sum_{x'=1}^{n} f(x')} \leq \frac{(10k)^{r+1}}{n}. 
		\end{equation*}
	\end{lemma}
	\begin{proof}
		By the pigeonhole principle, there is $f_i\in F$ and a set $X_i\subset \br{1,\cdots,n}$ of size at least $n /k$ such that for every $x\in X_i$ we have $f(x)=f_i(x)$; i.e., $f_i$ satisfies the set $X_i$.
		For every $x\in \br{1,\cdots,n}$ we have
		\begin{align}
			\label{general sen bound: eq1}
			\frac{f(x)}{\displaystyle \sum_{x' =1}^n f(x')} & \leq \frac{f_i(x)}{\displaystyle \sum_{x' =1}^n f(x')} \\
			\label{general sen bound: eq2}
			& \leq \frac{f_i(x)}{\displaystyle \sum_{x'\in X_i} f(x')} \\
			\label{general sen bound: eq3}
			& = \frac{f_i(x)}{\displaystyle \sum_{x'\in X_i} f_i(x')}\\
			\label{general sen bound: eq4}
			& \leq \frac{(10k)^{r+1}}{n},
		\end{align}
		where \eqref{general sen bound: eq1} is since $\displaystyle f(x) = \min_{f'\in F} f'(x)$ and $f_i\in F$, \eqref{general sen bound: eq2} is since $X_i\subset \br{1,\cdots,n}$, \eqref{general sen bound: eq3} is from the definition of $X_i$ as a set such that for every $x\in X_i$ we have $f(x)=f_i(x)$, and \eqref{general sen bound: eq4} is by substituting $X:=X_i$ and $\tilde{f}:=f_i$ in Lemma~\ref{general sen bound part 1}.
	\end{proof}
	\section{Deterministic coreset construction}\label{sec: det core}
	The following lemma is a modification of Lemma 11 from~\cite{k-Segmentation}, which is significantly influenced by it; we use constant sensitivity and move from sums to integrals.
	\begin{lemma}\label{lemma: deterministic construction}
		Let $k\geq 1$ and let $f:[0,1] \to [0,\infty)$ be a $k$-piece-wise monotonic function, where $\displaystyle t := \int_0^1 f(x) dx >0$.
		Let $s$ such that for every $x\in [0,1]$ we have $\displaystyle f(x) \leq  t s$.
		Put $\eps \in (0,1)$ and let $\displaystyle \eps' := \frac{1}{\big\lceil (2k s)/\eps\big\rceil}$.
		Let $S:= \br{i \cdot \eps' \mid i\in \br{0,\cdots,1/\eps'}}$.
		We have that 
		\[
		\abs{\frac{1}{|S|}\cdot \sum_{x\in S} f(x) -t } \leq \eps t.
		\]
	\end{lemma}
	\begin{proof}
		For every $i\in [0,1]$ let $h(i) := f(i)/(st)$.
		We will prove that
		\begin{equation} \label{eq: deterministic proof 0}
			\left| \int_0^1 h(x) dx -  \frac{1}{|S|}\cdot \sum_{x\in S} h(x) \right| \leq 2\eps' k.
		\end{equation}
		Multiplying this by $ts$ yields
		\begin{equation*}
			\left| \int_0^1 f(x) dx -  \frac{1}{|S|}\cdot \sum_{x\in S} f(x) \right| \leq 2\eps' k st \leq \eps t,
		\end{equation*}
		which proves the lemma.
		
		Since $f$ is $k$-piecewise monotonic, $h$ is $k$-monotonic. Hence, there is a partition
		$\Pi$ of $[0,1]$ into $k$ consecutive intervals such that $h$ is monotonic over each of these intervals.
		
		For every $j \in S$ let $b(j) := \lceil j /(\eps \eps' s)\rceil$ and $I_j:= \br{i\in [0,1] \mid b(i) = b(j)}$.
		Here, $b(j)$ assigns each sample $j\in S$ to a small sub-interval (bucket) of length $\eps \eps' s$.
		$I_j$ contains all points in $[0,1]$ in the same bucket. This ensures that $|h(x) - h(j)|$ is small for all $x \in I_j$.
		
		For every $I\in \Pi$ we define $G(I):=\br{j\in S \mid I_j \subset I}$.
		Their union is denoted by $\displaystyle G := \bigcup_{I \in \Pi} G(I)$.
		Hence,
		\begin{align}
			\label{eq: deterministic proof 0s}
			\Bigg| \int_0^1 h(x) dx -  \frac{1}{|S|}\cdot \sum_{x\in S} h(x) \Bigg| & = \left| \sum_{j\in S} \int_{x\in I_j} \Big( h(x) -  h(j) \Big) dx \right| \\
			& \leq \left| \sum_{j\in S \setminus G} \int_{x\in I_j} \Big(h(x) - h(j) \Big) dx \right| \label{eq: deterministic proof 1} \\
			& +\sum_{I\in \Pi} \left| \sum_{j\in G(I)} \int_{x\in I_j} \big( h(x) - h(j) \big) dx \right|. \label{eq: deterministic proof 2}
		\end{align}
		We now bound \eqref{eq: deterministic proof 1} and \eqref{eq: deterministic proof 2}.
		Pick $j\in S$. By the construction of $S$ in the lemma we have $|I_j\cup S| = 1$ and $\displaystyle \max\br{i\in I_j}- \min\br{i\in I_j}\leq \eps'$. Hence, 
		\begin{equation} \label{eq: deterministic proof 3}
			\left| \int_{x\in I_j} \Big( h(x) - h(j) \Big) dx \right| \leq \eps' \cdot \left( \max_{x\in I_j} h(x) - \min_{x\in I_j} h(x)\right) \leq \eps',
		\end{equation}
		where the second inequality holds since $h(x)\leq 1$ for every $x\in [0,1]$ (follows from the definition of $h$ and $s$).
		Since each set $I\in \Pi$ contains consecutive numbers, we have $|S\setminus G| \leq |\Pi| \leq k$. Using this and \eqref{eq: deterministic proof 3}, bounds \eqref{eq: deterministic proof 1} as follows
		\begin{equation}\label{eq: deterministic proof 3s}
			\left| \sum_{j\in S \setminus G} \int_{x\in I_j} \Big( h(x) - h(j) \Big) dx \right| \leq |S\setminus G| \cdot \eps' \leq k\eps'.
		\end{equation}
		Put $I\in \Pi$ and denote the numbers in $G(I)$ by $\br{g_1,\cdots,g_\theta}$.
		Recall that $h$ is monotonic over $I$.
		Without loss of generality, assume that $h$ is non-decreasing on $I$.
		Therefore, summing \eqref{eq: deterministic proof 2} over $G(I)$ yields
		\begin{align}
			\label{eq: deterministic proof 4.1}
			\Bigg| \sum_{j\in G(I)} \int_{x\in I_j} \Big( h(x) - h(j)  \Big)  dx \Bigg| 
			& \leq \sum_{j=g_1}^{g_\theta} \left|  \int_{x\in I_j} \Big( h(x) - h(j) \Big) dx \right|\\
			\label{eq: deterministic proof 4.2}
			& \leq \sum_{j=g_1}^{g_\theta} \eps' \cdot \left( \max_{x\in I_j} h(x) - \min_{x\in I_j} h(x)\right) \\
			\label{eq: deterministic proof 4.3}
			& \leq \eps' \cdot \sum_{j=g_1}^{g_\theta-1}  \left( \min_{x\in I_j} h(x) - \min_{x\in I_{j-1}} h(x) \right) \\
			\label{eq: deterministic proof 4.4}
			& = \eps' \cdot \left( \min_{x\in I_{g_\theta}} h(x) - \min_{x\in I_{g_1}} h(x)\right) \\
			\label{eq: deterministic proof 4.5}
			& \leq \eps',
		\end{align}
		where \eqref{eq: deterministic proof 4.4} is since the sum telescopes because the $I_j$ are consecutive and $h$ is monotone, and \eqref{eq: deterministic proof 4.5} is since $h(x) \leq 1$ for every $i\in [0,1]$.
		Summing over every $I\in \Pi$ bounds \eqref{eq: deterministic proof 2} as,
		\begin{equation}\label{eq: deterministic proof 5}
			\sum_{I\in \Pi} \left| \sum_{j\in G(I)} \int_{x\in I_j} \Big( h(x) - h(j) \Big) dx \right| \leq |\Pi| \cdot \eps \leq k\cdot \eps'.
		\end{equation}
		Plugging \eqref{eq: deterministic proof 5} and  \eqref{eq: deterministic proof 3s} in (\ref{eq: deterministic proof 0s}--\ref{eq: deterministic proof 2}) yields \eqref{eq: deterministic proof 0}, using $\eps' \le \eps/(2ks)$.
	\end{proof}
	\section{Analysis of Algorithm~\ref{algorithm - deterministic sample}}\label{sec: alg 1 proof}
	Using a similar method to the one in the proof of Lemma~\ref{general sen bound} yields the following sensitivity bound.
	\begin{lemma} \label{points to line: sen bound}
		Let $\Tilde{\ell}: [0,1]\to \REAL^d$ be a segment; see Definition~\ref{def: segment}.
		Let $n\geq 10k$.
		% and let $\displaystyle X=\br{ i/\Tilde{n} \mid i\in\br{1,\cdots,\Tilde{n}}}$.
		For every $x\in [0,1]$ and any weighted set $Q:=(P,w)$ of size $k$ we have
		\begin{equation*}
			\frac{\displaystyle \dist\big(Q,\Tilde{\ell}(x)\big)}{ \displaystyle 
				\sum_{i =1}^{n} \dist\big(Q,\Tilde{\ell}(i/n)\big)} \leq \frac{(20k)^{r+1}}{n}.
		\end{equation*}
	\end{lemma}
	\begin{proof}
		By the pigeonhole principle, there exists $p'\in P$ and a set \\$X\subset \br{i/n \mid i\in \br{1,\cdots,n}}$ of size at least $n/k$ such that for every $x\in X$ we have $\displaystyle p'\in \argmin_{p\in P} \lip(w(p)\cdot \dist(p,\Tilde{\ell}(x)))$.
		In other words, $X$ is the largest cluster, in terms of points assigned to $p'$.
		Let $u,v\in \REAL^d$ that defines $\Tilde{\ell}$ as in Definition~\ref{def: segment}.
		Let $\ell:\REAL \to \REAL^d$ such that for every $x'\in \REAL$ we have $\ell(x') = u+v\cdot x'$; i.e., an extension of $\Tilde{\ell}$ to a line.
		Let $\displaystyle \Tilde{x} \in \argmin_{x\in \REAL} w(p)\cdot \dist(p',\ell(x))$.
		Let $\lip_{p'}:[0,\infty) \to [0,\infty)$ where for every $\psi \in [0,\infty)$ we have $\lip_{p'}(\psi)=\lip(w(p')\cdot \psi)$.
		Since $\lip$ is $r$-log-Lipschitz (see Section~\ref{Problem definition} and Definition~\ref{def: log-Lipschitz}), so is $\lip_{p'}$.
		For every $i\in \br{1,\cdots,n}$ and $x= i/n$ we have
		\begin{align}
			\label{points to line: eq 1.1}
			\min_{p\in P} \lip(w(p)\cdot \dist(p,\ell(x)))
			& \leq \lip_{p'}(\dist(p',\ell(x))) \\
			\label{points to line: eq 1.2}
			& \leq \lip_{p'}(\dist(p',\ell(\Tilde{x}))+ \dist(\ell(x),\ell(\Tilde{x})) \\ 
			\label{points to line: eq 1.3}
			& \leq 2^{r} \cdot \lip_{p'}(\dist(p',\ell(\Tilde{x})) 
			+ 2^{r} \cdot \lip_{p'}(\dist(\ell(x),\ell(\Tilde{x}))),
		\end{align}
		where \eqref{points to line: eq 1.1} is since the minimum is over $P$ and we have $p'\in P$, \eqref{points to line: eq 1.2} is by the triangle inequality (for Euclidean distance), the \eqref{points to line: eq 1.3} is by property (2.3) of Lemma 2.1 in \cite{outliers-resistance}, where substituting $M:=\REAL^d$, $r:=r$, $\mathrm{dist}(p,q):=\dist(p,q)$ for every $p,q\in  \REAL^d$, and $\dist(x):=\lip_{p'}(x)$ for every $x\in [0,\infty)$.
		
		Observe that $\ell(\Tilde{x})$ is the projection of $p$ on the line $\ell$.
		Hence, for every $x'\in X$, by looking on the right angle triangle defined by $\ell(x'),\ell(\Tilde{x}),p$, we have
		\begin{equation} \label{points to line: eq 2}
			\dist(p',\ell(x')) \geq \dist(p',\ell(\Tilde{x})),\dist(\ell(x'),\ell(\Tilde{x})).
		\end{equation}
		Hence,
		\begin{align}
			\label{points to line: eq 3.0}
			\frac{\displaystyle \dist\big(Q,\Tilde{\ell}(x)\big)}{ \displaystyle 
				\sum_{i =1}^{n} \dist\big(Q,\Tilde{\ell}(i/n)\big)} & = \frac{\displaystyle \min_{p\in P}  \lip(w(p)\dist(p,\ell(x)))}{\displaystyle \sum_{x'=1}^n \min_{p\in P} \lip(w(p) \dist(p,\ell(\Tilde{x}/n)))} \\
			\label{points to line: eq 3.1}
			&  \leq 
			2^{r} \cdot \frac{\lip_{p'}(\dist(p',\ell(\Tilde{x}))) + \lip_{p'}(\dist(\ell(x),\ell(\Tilde{x})))}{\displaystyle \sum_{x'\in X} \lip_{p'}(\dist(p',\ell(x')))}\\
			\label{points to line: eq 3.2}
			& \leq \frac{2^{r}}{n} + 2^{r}\cdot \frac{\lip_{p'}(\dist(\ell(x),\ell(\Tilde{x})))}{\displaystyle \sum_{x'\in X} \lip_{p'}(\dist(\ell(\Tilde{x}),\ell(x')))},
		\end{align}
		where \eqref{points to line: eq 3.0} is by $Q=(P,w)$ and the definition of $\dist$, \eqref{points to line: eq 3.1} is by (\ref{points to line: eq 1.1}--\ref{points to line: eq 1.3}), and \eqref{points to line: eq 3.2} is by \eqref{points to line: eq 2}.
		
		Observe that for every $a,b\in \REAL$ we have
		\begin{align}
			\label{points to line: eq 4.1.1}
			\dist(\ell(a),\ell(b)) & = \dist(v\cdot a+u,v\cdot b+u) \\
			\label{points to line: eq 4.1.2}
			& = \| v\cdot (a-b) \|_2 \\
			\label{points to line: eq 4.1.3}
			& = \|v\|_2 \cdot |a-b|, 
		\end{align}
		where \eqref{points to line: eq 4.1.1} is by recalling the definition of $\ell$ as a function that for every $x\in \REAL$ returns $u+x\cdot v$, \eqref{points to line: eq 4.1.2} is since $\dist$ is the Euclidean distance function, and \eqref{points to line: eq 4.1.3} is by norm-2 properties.
		
		Let $\Tilde{f}:\REAL\to [0,\infty)$ such that for every $x'\in\REAL$ we have $\Tilde{f}(x')=\lip_{p'}(\|v\|_2 \cdot |x'-\Tilde{x}|)$.
		Since $\lip_{p'}$ is an $r$-log-Lipschitz function we have that $\Tilde{f}$ is a symmetric-$r$ function; see Definition~\ref{def: symmetric-$r$ function}.
		Hence,
		\begin{align}
			\label{points to line: eq 4.1}
			\frac{\lip_{p'}(\dist(\ell(x),\ell(\Tilde{x})))}{\displaystyle \sum_{x'\in X} \lip_{p'}(\dist(\ell(\Tilde{x}),\ell(x')))} & = 
			\frac{\lip_{p'}(\|v\|_2\cdot |x-\Tilde{x}|)}{\displaystyle \sum_{x'\in X} \lip_{p'}(\|v\|_2\cdot|x'-\Tilde{x}|)} \\  
			\label{points to line: eq 4.2}
			& = \frac{\Tilde{f}(x)}{\displaystyle \sum_{x'\in X} \Tilde{f}(x')}\\
			\label{points to line: eq 4.3}
			& \leq \frac{(10k)^{r+1}}{n},
		\end{align}
		where \eqref{points to line: eq 4.1} is by assigning (\ref{points to line: eq 4.1.1}--\ref{points to line: eq 4.1.3}), \eqref{points to line: eq 4.2} is by assigning the definition of the function $\tilde{f}$, and \eqref{points to line: eq 4.3} is by assigning $X$ and $\tilde{f}$ in Lemma~\ref{general sen bound part 1}; recall that $\tilde{f}$ is a symmetric-$r$- function, and that $|X|\subseteq \br{1,\cdots,n}$ is a set of size at least $n/k$.
		
		Combining (\ref{points to line: eq 3.0}--\ref{points to line: eq 3.2}) and (\ref{points to line: eq 4.1}--\ref{points to line: eq 4.3}) proves the lemma.
	\end{proof}
	Taking $n$ to infinity and using Riemann integration yields the following lemma.
	\begin{lemma}\label{points to line: sen bound 2}
		Let $\ell:[0,1]\to \REAL^d$ be a segment; see Definition~\ref{def: segment}.
		For every weighted set $Q$ of size $k$ and any $x'\in [0,1]$ we have
		\begin{equation*}
			\dist\big( Q,\ell(x') \big) \leq (20k)^{r+1} \cdot \int_0^1 \dist\big( Q,\ell(x) \big) dx.
		\end{equation*}
	\end{lemma}
	\begin{proof}
		From the definition of the Riemann integral, for every $x'\in [0,1]$ we have
		\begin{multline*}
			\dist\big( Q,\ell(x') \big) \leq (20k)^{r+1} \cdot \lim_{n\in \mathbb{Z},n \to \infty} \frac{1}{n} \cdot \sum_{i=1}^{n} 
			\dist\big( Q,\ell(x/n) \big)\\
			=  (20k)^{r+1} \cdot \int_0^1 
			\dist\big( Q,\ell(x) \big) dx,
		\end{multline*}
		where the inequality is by assigning $\Tilde{\ell}:=\ell$ and $n\to \infty$ in Lemma~\ref{points to line: sen bound}.
	\end{proof}
	
	The assignment of this sensitivity bound in Lemma~\ref{lemma: deterministic construction} requires a bound on the number of monotonic functions, which is presented in the following observation.
	
	\begin{lemma} \label{ob: minimum is 2k-monotonic}
		Let $\ell:[0,1]\to \REAL^d$ be a segment; see Definition~\ref{def: segment}.
		Let $(C,w)$ be weighted set of size $|C|=k$ and $f:[0,1] \to [0,\infty)$ such that for every $x\in [0,1]$ we have $\displaystyle f(x) := \dist\big((C,w),\ell(x)\big)$.
		It holds that $f$ is $2k$-piece-wise monotonic.
	\end{lemma}
	\begin{proof}
		Observe that the weighted Voronoi diagram for the weighted set $(C,w)$ (using $\dist(p,p')$, the Euclidean distance, as the distance between any two points $p,p\in \REAL^d$) has $k$ convex cells, and let $C'$ be the set of those cells.
		By the definition of the Voronoi diagram, for each cell in $C'$, that corresponds to the point $p\in C$, and for every $x\in [0,1]$, such that $\ell(x)$ is in the cell, we have 
		\[ 
		\min_{p'\in P} w(p) \dist(p',\ell(x)) = w(p) \dist(p,\ell(x)).
		\]
		Let $g: [0,1]\to [0,\infty)$ such that for every $x\in[0,1]$ we have 
		\begin{equation}\label{eq: minimum is 2k-monotonic 2}
			g(x) := \min_{p'\in P} w(p') \dist(p',\ell(x)),
		\end{equation}
		i.e., $f(x)$ without applying the $r$-Lipschitz function $\lip$ over the Euclidean distance.
		We have that $g$ has at most one extremum inside each cell in $C'$ (the single local minimum) and besides this only when transitioning between cells in $C'$, which, due to the cells being convex shapes, happens at most $k-1$ times.
		Therefore, $g$ has at most $2k-1$ extrema, and as such is $2k$-piece-wise monotonic.
		
		Let $x\in [0,1]$, and let 
		\[
		p \in \argmin_{p'\in P} w(p') \dist(p',\ell(x)).
		\]
		Hence, for every $p'\in P$ it holds that
		\[
		w(p) \dist(p,\ell(x)) \leq w(p') \dist(p',\ell(x)).
		\]
		Therefore, since $\lip:\REAL\to \REAL$ is an $r$-Lipschitz and as such non-decreasing, for every $p'\in P$ we have
		\begin{equation}\label{eq: minimum is 2k-monotonic 3}
			\lip\big(w(p) \dist(p,\ell(x))\big) \leq \lip\big(w(p') \dist(p',\ell(x))\big).
		\end{equation}
		Combining this with the definition of the functions $f$ and $g$ yields
		\begin{align}
			\label{eq: minimum is 2k-monotonic 4.1}
			f(x) & =  \min_{p'\in P} \lip\big(w(p') \dist(p',\ell(x))\big) \\
			\label{eq: minimum is 2k-monotonic 4.2}
			& = \lip\big(w(p) \dist(p,\ell(x))\big) \\
			\label{eq: minimum is 2k-monotonic 4.3}
			& = \lip \big(\min_{p'\in P} w(p) \dist(p',\ell(x)) \big) \\
			\label{eq: minimum is 2k-monotonic 4.4}
			& = \lip(g(x)),
		\end{align}
		where Equation~\ref{eq: minimum is 2k-monotonic 4.1} is by the definition of $f$, Equation~\ref{eq: minimum is 2k-monotonic 4.2} is by \eqref{eq: minimum is 2k-monotonic 3}, \eqref{eq: minimum is 2k-monotonic 4.3} is by assigning that $\lip:\REAL\to \REAL$ is a $r$-Lipschitz and as such non-decreasing, and Equation~\ref{eq: minimum is 2k-monotonic 4.4} is by plugging the definition of $g:[0,1]\to [0,\infty)$ from Equation~\ref{eq: minimum is 2k-monotonic 2}.
		
		Hence, for every $x\in [0,1]$ we have
		\[
		f(x) = \lip(g(x)).
		\]
		Since $g$ is a $2k$-piece-wise monotonic function, and due to $lip$ being a non-decreasing function, we have that $f$ is a $2k$-piece-wise monotonic function; applying $\lip$ over any increasing segment of $g$ would keep the segment increasing, and the same would hold for decreasing segments.
	\end{proof}
	
	\begin{theorem}\label{deterministic sample theorem proof}
		Let $\ell: [0,1] \to \REAL^d$ be a segment, and let $\eps \in (0,1/10]$; see Definition~\ref{def: segment}.
		Let $(S,w)$ be the output of a call to $\OneSegmentCoreset(\ell,k,\eps)$; see Algorithm~\ref{algorithm - deterministic sample}. 
		Then $(S,w)$ is an $(\eps,k)$-coreset for $\ell$; see Definition~\ref{def: eps coreset}.
	\end{theorem}
	\begin{proof}
		Let $Q$ be a weighted set of size $|Q|=k$ and $f:[0,1] \to [0,\infty)$ such that for every $x\in[0,1]$ we have $\displaystyle f(x) = \dist\big(Q, \ell(x) \big)$.
		For every $x'\in[0,1]$, by assigning $x',\ell$, and $Q$ in Lemma~\ref{points to line: sen bound 2}, we have
		\begin{equation*}
			f(x') \leq (20k)^{r+1} \cdot \int_0^1 f(x) dx.
		\end{equation*}
		If $\displaystyle \int_0^1 f\big(x\big) dx =0$, the lemma holds from the construction of Algorithm~\ref{algorithm - deterministic sample}, hence, we assume this is not the case.
		By Observation~\ref{ob: minimum is 2k-monotonic} $f$ is a $2k$-piece-wise monotonic.
		Hence, assigning $\eps,f$, and substituting $s := (20k)^{r+1}, k:=2k$ in Lemma~\ref{lemma: deterministic construction} combined with the construction of Algorithm~\ref{algorithm - deterministic sample} proves the lemma.
	\end{proof}
	\section{Proof of the desired properties of Algorithm~\ref{algorithm - coreset}}
	In this section, we prove the desired properties of Algorithm~\ref{algorithm - coreset}, as previously stated in Theorem~\ref{coreset theorem}.
	\begin{theorem} \label{coreset theorem proof}
		Let $L$ be a set of $n$ segments, and let $\eps,\delta \in (0,1/10]$.
		Let $\displaystyle m:=\frac{8k n \cdot (20k)^{r+1}}{\eps}$ and $k'\in (k+1)^{O(k)}$.
		Let $(P,w)$ be the output of a call to $\Coreset(L,k,\eps,\delta)$; see Algorithm~\ref{algorithm - coreset}.
		Then Claims~(i)--(iii) hold as follows:
		\begin{enumerate}[(i)]
			\item With probability at least $1-\delta$, we have that $(P,w)$ is an $(\eps,k)$-coreset of $L$; see Definition~\ref{def: eps coreset}.
			\item $\displaystyle |P| \in \frac{k' \cdot \log^2 m }{\eps^2} \cdot O\of{d^* +\log\of{\frac{1}{\delta}}}$.
			\item The $(\eps,k)$-coreset $(P,w)$ can be computed in time
			\[
			m t k' + t k' \log(m) \cdot \log^2\big(\log(m)/\delta\big) 
			+ \frac{k' \log^3 m}{\eps^2} \cdot \left(d^* + \log \of{\frac{1}{\delta}} \right).
			\]
		\end{enumerate}
	\end{theorem}
	\begin{proof}
		Properties (ii) and (iii) follow from the construction of Algorithm~\ref{algorithm - coreset}, and the corresponding Claims (ii) and (iii) in Theorem~\ref{th: outliers-resistance coreset}.
		Hence, it suffices to prove Claim (i).
		
		Let $P'\subset \REAL^d$ and $\eps'>0$ as computed in the call to $\Coreset(L,\eps,\delta)$.
		Let $\psi:P'\to [0,\infty)$ such that for every $p\in P$ we have $\psi(p):=1/\eps'$.
		By Theorem~\ref{deterministic sample theorem} and the construction of Algorithm~\ref{algorithm - coreset} we have that $(P',\psi)$ is an $(\eps/2)$-coreset for $L$; see Definition~\ref{def: eps coreset}.
		By Theorem~\ref{th: outliers-resistance coreset}, more specifically its Claim (i), and the construction of Algorithm~\ref{algorithm - coreset}, with probability at least $1-\delta$, for every $Q$, a weighted set of size $|Q|=k$, we have
		\begin{equation}\label{eq: coreset theorem proof 1}
			\left| \sum_{p\in P} w(p) \dist(Q,p) - \frac{1}{\eps'} \sum_{p\in P'} \dist(Q,p) \right| \leq
			\frac{\eps}{4}\sum_{p\in S} w(p) \dist(Q,p).
		\end{equation}
		Suppose this indeed occurs.
		For every $Q$, a weighted set of size $|Q|=k$, we have
		\begin{align}
			\label{eq: coreset theorem proof 2}
			\Bigg| \loss(L,Q) -  \sum_{p\in P}  w(p)\dist(Q,p)\Bigg| & \leq \left| \loss(L,Q) - \sum_{p\in P'} \psi(p) \dist(Q,p)\right|\\
			\label{eq: coreset theorem proof 2.5}
			+ & \left|  \sum_{p\in P} w(p) \cdot \dist(Q,p) - \sum_{p\in P'} \psi(p) \dist(Q,p)\right| \\
			\label{eq: coreset theorem proof 3}
			& \leq \frac{\eps}{2} \loss(L,Q)  + \frac{\eps}{4} \sum_{p\in P'} \psi(p)\dist(Q,p)\\
			\label{eq: coreset theorem proof 4}
			\leq \; &
			\frac{\eps}{2}  \loss(L,Q) + \frac{\eps}{4 (1-\eps/2)} \loss(L,Q) \\
			\label{eq: coreset theorem proof 5}
			& < \eps \cdot \loss(L,Q),
		\end{align}
		where \eqref{eq: coreset theorem proof 2}--\eqref{eq: coreset theorem proof 2.5} is by the triangle inequality, \eqref{eq: coreset theorem proof 3} is since  $(P',\psi)$ is an $(\eps/2)$-coreset for $L$ and plugging \eqref{eq: coreset theorem proof 1}, \eqref{eq: coreset theorem proof 4} is since $(P',\psi)$ is an $(\eps/2)$-coreset for $L$, and \eqref{eq: coreset theorem proof 5} is by assigning that $\eps\in(0,1/10]$, hence, $(1-\eps) > 1/2$.
		Thus, with probability at least $1-\delta$, $(P,w)$ is an $\eps$-coreset for $L$.
	\end{proof}
	
	\section{Generalization to multi-dimensional shapes.} \label{sec: Algorithm 3}
	Having established coreset constructions for segments, we now generalize to arbitrary well-bounded convex sets.
	The linear structure of segments simplifies analysis, which we extend to convex sets via volume-based sampling.
	Through this section, let $d\geq 2$.
	Recall the definition of $d^*,r$, and $\dist$ from Section~\ref{Problem definition}.
	
	This generalization requires the following definitions.
	\begin{definition} [loss-function] \label{def: convex loss}
		Let $\mathcal{A} \subset \REAL^d$ be a convex set.
		We define the loss of fitting every weighted set $Q:=(C,w)$ of size $|C|=k$, as
		$\displaystyle
		\loss\big(Q,\mathcal{A}\big) := 
		\int_{p\in \mathcal{A}} \dist(Q,p) d V,
		$
		where $dV$ is the volume element.
		Given a finite set $\mathcal{K}$ of convex sets, and a weighted set $Q:=(C,w)$ of size $|C|=k$, we define the loss of fitting $Q$ to $\mathcal{K}$ as
		$\displaystyle
		\loss(Q,\mathcal{K}) := \sum_{\mathcal{A} \in \mathcal{K}} \loss(Q,\mathcal{A}).
		$
		Given such a set $\mathcal{K}$, the goal is to recover a weighted set $Q:=(C,w)$ of size $|C|=k$, that minimizes $\loss\big(Q,\mathcal{K}\big)$.
	\end{definition}
	
	\begin{definition} [convex-$(\eps,k)$-coreset] \label{def: convex eps coreset}
		Let $\mathcal{K}$ be a set of $n$ convex sets in $\REAL^d$, and let $\eps>0$ be an error parameter; see Definition~\ref{def: segment}.
		A weighted set $(S,w)$ is a \emph{convex-$(\eps,k)$-coreset} for $\mathcal{K}$ if for every weighted set $Q:=(C,w)$ of size $|C|=k$, we have 
		\[
		\abs{\loss(Q,\mathcal{K}) - \sum_{p\in S} w(p)\cdot \dist(Q,p)} \leq
		\eps \cdot \loss(Q,\mathcal{K}).
		\]
	\end{definition}

	\begin{definition} [well-bounded set] \label{def: well bounded set}
		A convex set $\mathcal{A}\subset \REAL^d$ is \emph{well-bounded} if there is an ``oracle membership'' function $\psi: \REAL^d\to \br{0,1}$ that maps every $p\in \mathcal{A}$ to $1$ and $p\in \REAL^d \setminus \mathcal{A}$ to $0$, where the  output can be computed in $d^{O(1)}$ time.
		Furthermore, a minimum-volume bounding box (not necessarily axis-aligned) enclosing 
		$\mathcal{A}$ is provided, with non-zero volume.
	\end{definition}
	This definition attempts to define ``not-over complex'' shapes and contains hyper-rectangles, spheres, and ellipsoids.
	
	As in Algorithm~\ref{algorithm - coreset}, the intermediate sampling at Algorithm~\ref{algorithm - convex coreset} is a point set of size larger than the original input size, which is then reduced to a point set of size that is poly-logarithmic in the size of the input set.
	\clearpage
	\begin{algorithm}[h]
		\caption{$\ConvexCoreset(\mathcal{K},k,\eps,\delta)$;\\
			see Theorem~\ref{convex coreset theorem}.\label{algorithm - convex coreset}}
		\SetKwInOut{Input}{Input}
		\SetKwInOut{Output}{Output}
		
		\Input{A set $\mathcal{K}$ of $n$ well-bounded convex sets, an integer $k\geq 1$, and $\eps,\delta\in (0,1/10]$.}
		\Output{A weighted set $(P,w)$, which, with probability at least $1-\delta$ is a convex-$(\eps,k)$-coreset for $\mathcal{K}$; see Definition~\ref{def: convex eps coreset}.}
		
		Set $t:= (20k)^{d(r+1)}$.
		
		Set $\displaystyle \lambda:=  \frac{c^* d^* (t+1)}{\eps^2} \left(k \log(t+1) + \log\left( \frac{2}{\delta}\right)\right)$, where $c^*\geq 1$ is a constant that can be determined from the proof of Theorem~\ref{convex coreset theorem}.
		
		\For{every $\mathcal{A}\in \mathcal{K}$}
		{
			Let $B_{\mathcal{A}}\subset \REAL^d$ be a minimum-volume bounding box (not necessarily axis-aligned) enclosing $\mathcal{A}$.
			
			$S_{\mathcal{A}} := \emptyset$
			
			\While{$|S_{\mathcal{A}}|< \lambda$}
			{
				Let $p\in B_{\mathcal{A}}$ be sampled uniformly at random from $B_{\mathcal{A}}$.
				
				\If{$p\in \mathcal{A}$}
				{
					$S_{\mathcal{A}}:=S_{\mathcal{A}} \cup p$
				}
			}
			
		}
		
		$\displaystyle S:= \bigcup_{\mathcal{A}\in \mathcal{K}} S_{\mathcal{A}}$.
		
		$(P,w'):=\textsc{Core-set}(S,k,\eps/4,\delta/2)$; see Definition~\ref{def: global-lip}.
		
		Set $w(p) := \lambda w'(p)$ for every $p\in P$.
		
		\Return $(P,w)$.
	\end{algorithm}
	
	The following theorem states and proves the desired properties of Algorithm~\ref{algorithm - convex coreset}.
	\begin{theorem}\label{convex coreset theorem}
		Let $\mathcal{K}$ be a set of $n\geq 1$ well-bounded convex sets; see Definition~\ref{def: well bounded set}.
		Put $\eps,\delta\in(0,1/10]$.
		Let $(P,w)$ be the output of a call to\\ $\ConvexCoreset(\mathcal{K},k,\eps,\delta)$; see Algorithm~\ref{algorithm - convex coreset}.
		Let $\lambda$ be as computed at Line 2 at the call to Algorithm~\ref{algorithm - convex coreset}.
		Let $n':=\lambda n$ and $t:=(20k)^{d(r+1)}$.
		Then there is $k' \in (k+1)^{O(k)}$ such that Claims~(i)--(iii) hold as follows:
		\begin{enumerate}[(i)]
			\item With probability at least $1-\delta$ we have that $(P,w)$ is a convex-$(\eps,k)$-coreset of $\mathcal{K}$; see Definition~\ref{def: convex eps coreset}.
			\item The size of the set $P$ is in
			$\displaystyle \frac{k' \cdot \log^2 n' }{\eps^2} \cdot O\of{d^* +\log\of{\frac{1}{\delta}}}.$
			\item The expected time to compute $(P,w)$ is in 
			\[
			n'd^{O(1)}+ n' t k' + t k' \log(n') \cdot \log^2\big(\log(n')/\delta\big) 
			+ \frac{k' \log^3 n'}{\eps^2} \cdot \left(d^* + \log \of{\frac{1}{\delta}} \right).
			\]
		\end{enumerate}
	\end{theorem}
	\begin{proof}
		Property (ii) follows from the construction of Algorithm~\ref{algorithm - convex coreset}, and the corresponding Claim (ii) in Theorem~\ref{th: outliers-resistance coreset}.
		Hence, we will prove the only remaining claims, Claims (i) and (iii) of the theorem.
		\clearpage
		
		\textbf{Proof of Claim (i): correctness.}
		Let $S\subset \REAL^d$ and $\lambda>0$ as computed in the call to $\ConvexCoreset(\mathcal{K},\eps,\delta)$.
		Let $\psi:S\to [0,\infty)$ such that for every $p\in S$ we have $\psi(p)=\lambda$.
		Let $p\in \mathcal{A}$, which is in a segment $\ell:[0,1]\to \REAL^d$ that is in $\mathcal{A}$, by Lemma~\ref{points to line: sen bound}, for every weighted set $Q$ of size $k$, any $n\geq 10 k$, and all $x\in[0,1]$ we have
		\[
		\frac{\displaystyle \dist\big(Q,\ell(x)\big)}{ \displaystyle 
			\sum_{i =1}^{n} \dist\big(Q,\ell(i/n)\big)} \leq \frac{(20k)^{r+1}}{n}.
		\]
		Hence, by taking $n$ to infinity by Riemann integrals, we have 
		\[
		\dist\big(Q,\ell(x)\big) \leq (20k)^{r+1} \int_0^1 \dist\big(Q,\ell(y)\big) dy.
		\]
		Therefore, since a convex set can be considered as an infinite union of segments, repeating the claim above $d$ times yields that for every weighted set $Q$ of size $k$, for every $p\in C$ we have
		\begin{equation}\label{eq: convex sen bound}
			\dist\big(Q,p\big)\leq (20k)^{d(r+1)} \int_{p\in C}\dist\big(Q,p\big) dV.
		\end{equation}
		Let $\mathcal{A}\in \mathcal{K}$, with the corresponding bounding box $B_C$.
		Substituting $\eps:=\eps,\delta:=\delta,\lambda:=\lambda$, the query space $\left(P,\Q,F,\|\cdot \|_1\right)$, where $\Q$ is the union over all the weighted sets of size $k$, $d^*$ the VC-dimension induced by $\Q$ and $F$ from Definition~\ref{def: global-lip}, the sensitivity bound $\displaystyle \tilde{s}:= \frac{(20k)^{d(r+1)}}{n'}$ for every sufficiently large $n$ (follows from Equation~\ref{eq: convex sen bound}), and the total sensitivity $\displaystyle t = (20k)^{d(r+1)}$ that follows from Equation~\ref{eq: convex sen bound}, yields that with probability at least $1-\delta/(n+1)$, a uniform sample (with appropriate weights) of size $\lambda$, as defined at Line 1 of Algorithm~\ref{algorithm - convex coreset}, is a convex-$(\eps/2)$-coreset for $\mathcal{A}$.
		Thus, by the construction of $S_{\mathcal{A}}$ at $B_{\mathcal{A}}$ it follows that $(S_{\mathcal{A}},\psi)$ is with probability at least $1-\delta/(n+1)$ a convex-$(\eps/2)$-coreset for $\mathcal{A}$.
		
		Suppose that this is indeed the case that for every $\mathcal{A}\in \mathcal{K}$ we have that $(S_{\mathcal{A}},\psi)$ is a convex-$(\eps/2)$-coreset for $\br{\mathcal{A}}$, which occurs, by the union bound, with probability at least $\displaystyle 1 - \frac{n \delta}{n+1}$. 
		Hence, by its construction at Algorithm~\ref{algorithm - convex coreset}, $(S,\psi)$ is a convex-$(\eps/2)$-coreset for $\mathcal{K}$.
		
		By Theorem~\ref{th: outliers-resistance coreset}, more specifically its Claim (i), and the construction of Algorithm~\ref{algorithm - convex coreset}, with probability at least $1-\delta/(n+1)$, for every $Q$ a weighted set of size $|Q|=k$, we have
		\begin{equation}\label{eq: convex coreset theorem proof 1}
			\left| \sum_{p\in S} w(p) \cdot \dist(Q,p) - \lambda \cdot \sum_{p\in P'} \dist(Q,p) \right| 
			\leq
			\frac{\eps}{4}\cdot \sum_{p\in S} w(p) \cdot \dist(Q,p).
		\end{equation}
		Suppose this indeed occurs.
		For every $Q$, a weighted set of size $|Q|=k$, we have
		\begin{align}
			\Bigg| \loss(\mathcal{K},Q) - \sum_{p\in S}  w(p)  \cdot \dist(Q,p)\Bigg| & \nonumber\\
			\label{eq: convex coreset theorem proof 2}
			& \leq \left| \loss(\mathcal{K},Q) - \sum_{p\in P'} \psi(p) \cdot \dist(Q,p)\right| \\
			\label{eq: convex coreset theorem proof 2.5}
			+ & \left| \sum_{p\in S} w(p) \cdot \dist(Q,p) - \sum_{p\in P'} \psi(p) \cdot \dist(Q,p)\right| \\
			\label{eq: convex coreset theorem proof 3}
			& \leq \frac{\eps}{2} \cdot \loss(\mathcal{K},Q)  + \frac{\eps}{4} \cdot \sum_{p\in P'} \psi(p) \cdot \dist(Q,p)\\
			\label{eq: convex coreset theorem proof 4}
			& \leq
			\frac{\eps}{2} \cdot \loss(\mathcal{K},Q) + \frac{\eps}{4 (1-\eps)} \cdot \loss(\mathcal{K},Q) \\
			\label{eq: convex coreset theorem proof 5}
			& < \eps \cdot \loss(\mathcal{K},Q),
		\end{align}
		where \eqref{eq: convex coreset theorem proof 2}--\eqref{eq: convex coreset theorem proof 2.5} is by the triangle inequality, \eqref{eq: convex coreset theorem proof 3} is since  $(S,\psi)$ is a convex-$(\eps/2)$-coreset for $\mathcal{K}$ and \eqref{eq: convex coreset theorem proof 1}, \eqref{eq: convex coreset theorem proof 4} is since $(S,\psi)$ is a convex-$(\eps/2)$-coreset for $\mathcal{K}$, and \eqref{eq: convex coreset theorem proof 5} is by assigning that $\eps\in(0,1/10]$, hence, $(1-\eps) > 1/2$.
		Thus, by the union bound, with probability at least 
		\[
		\Big(1-\delta n/(n+1)\Big)\big(1-\delta/(n+1)\big)\leq 1-\delta (n+1)/(n+1) = 1-\delta,
		\]
		$(S,w)$ is an convex-$\eps$-coreset for $\mathcal{K}$.
		
		\textbf{Proof of Claim (iii): expected running time.}
		Since every $\mathcal{A}\in \mathcal{K}$ is a well-bounded convex shape, with a given bounding box, the running time of Lines [1--8] of the call to $\ConvexCoreset(\mathcal{K},k,\eps,\delta)$ is in $d^{O(1)}$ times order of the number of iterations in the ``while" loop at Line 5 of Algorithm~\ref{algorithm - convex coreset}.
		
		Since every $\mathcal{A}\in \mathcal{K}$ is a convex set, by the properties of John ellipsoid it follows that each iteration of the innermost ``while'' loop, at Line 5 of Algorithm~\ref{algorithm - convex coreset}, adds with probability at least $1/d^2$ a point to $S_{\mathcal{A}}$; follows from observing that the John ellipsoid $E_{\mathcal{A}}$ bounding $\mathcal{A}$ has volume at most $d$ of $C$ (i.e. $d \cdot \mathrm{Vol}(\mathcal{A})$) and the bounding box $B_{\mathcal{A}}$ bounds $E_{\mathcal{A}}$ and has a john ellipsoid with volume $d$ times the volume of $E_{\mathcal{A}}$.
		
		Thus, the expected number of iterations in the innermost "while" loop at Line 5 of Algorithm~\ref{algorithm - convex coreset} is in $O(n d^2 \lambda)$.
		
		Hence, the expected running time of Lines [1--8] of the call to\\ $\ConvexCoreset(\mathcal{K},k,\eps,\delta)$ is in $d^{O(1)} n\lambda$.
		As such, combining this with Claim (iii) of Theorem~\ref{th: outliers-resistance coreset} yields the desired expected running time stated in Claim (iii) of the theorem.
	\end{proof}

\end{document}